\documentclass[10pt,twocolumn,letterpaper]{article}

\usepackage[pagenumbers]{cvpr} 

\usepackage[dvipsnames]{xcolor}

\definecolor{cvprblue}{rgb}{0.21,0.49,0.74}
\usepackage[pagebackref,breaklinks,colorlinks,citecolor=cvprblue]{hyperref}
\usepackage{bm}

\newtheorem{proposition}{Proposition}

\newtheorem{remark}{Remark}

\newtheorem{subproposition}{Part}[proposition]

\newcommand{\sign}{\text{sign}}

\newcommand{\softmax}{\text{softmax}}
\newcommand{\sg}{\text{sg}}

\newcommand{\mystar}{\raisebox{0.5ex}{\tiny *}}

\usepackage{subcaption}

\usepackage{multirow} 
\usepackage{makecell}

\usepackage{xcolor}
\usepackage{tikz}

\definecolor{myred}{rgb}{1.0, 0.62, 0.61}
\definecolor{mygreen}{RGB}{8, 147, 146}
\definecolor{myorange}{RGB}{233, 159, 105}
\definecolor{mypink}{RGB}{207, 89, 126}

\newcommand{\cb}[1]{
    \hspace{-0.5em}
  \begin{tikzpicture}[baseline=-0.35em]
  \pgfmathsetmacro{\opacitylevel}{#1/100}
    \node[circle, fill=myred, fill opacity=\opacitylevel, inner sep=0.2em] at (0,0) {};
  \end{tikzpicture}
  \hspace{-0.6em}
}

\usepackage{setspace}
\usepackage{tocloft}

\usepackage[accsupp]{axessibility}

\def\paperID{6381} 
\def\confName{CVPR}
\def\confYear{2024}

\begin{document}
\addtocontents{toc}{\protect\setcounter{tocdepth}{-1}}
\title{Neural Lineage}

\author{\bf Runpeng Yu \quad 
Xinchao Wang$^{\dagger}$\\
National University of Singapore\\
{\tt\small r.yu@u.nus.edu} \quad {\tt\small xinchao@nus.edu.sg}
}

\twocolumn[{%
\renewcommand\twocolumn[1][]{#1}%
\maketitle
\vspace{-2.5em}
\begin{center}
    \centering
    \captionsetup{type=figure}
    \begin{subfigure}[b]{0.18\linewidth}
        \includegraphics[width=\linewidth]{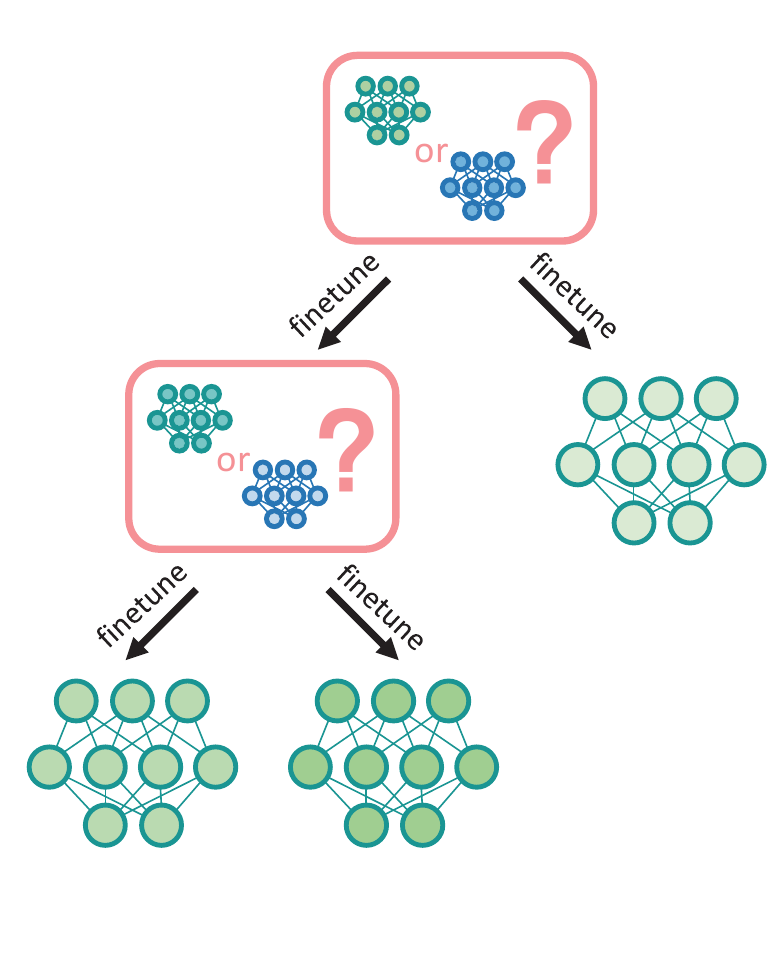}
        \caption{Task illustration}
        \label{fig:head:task}
    \end{subfigure}
    \hfill
    \begin{subfigure}[b]{0.4\linewidth}
        \includegraphics[width=\linewidth]{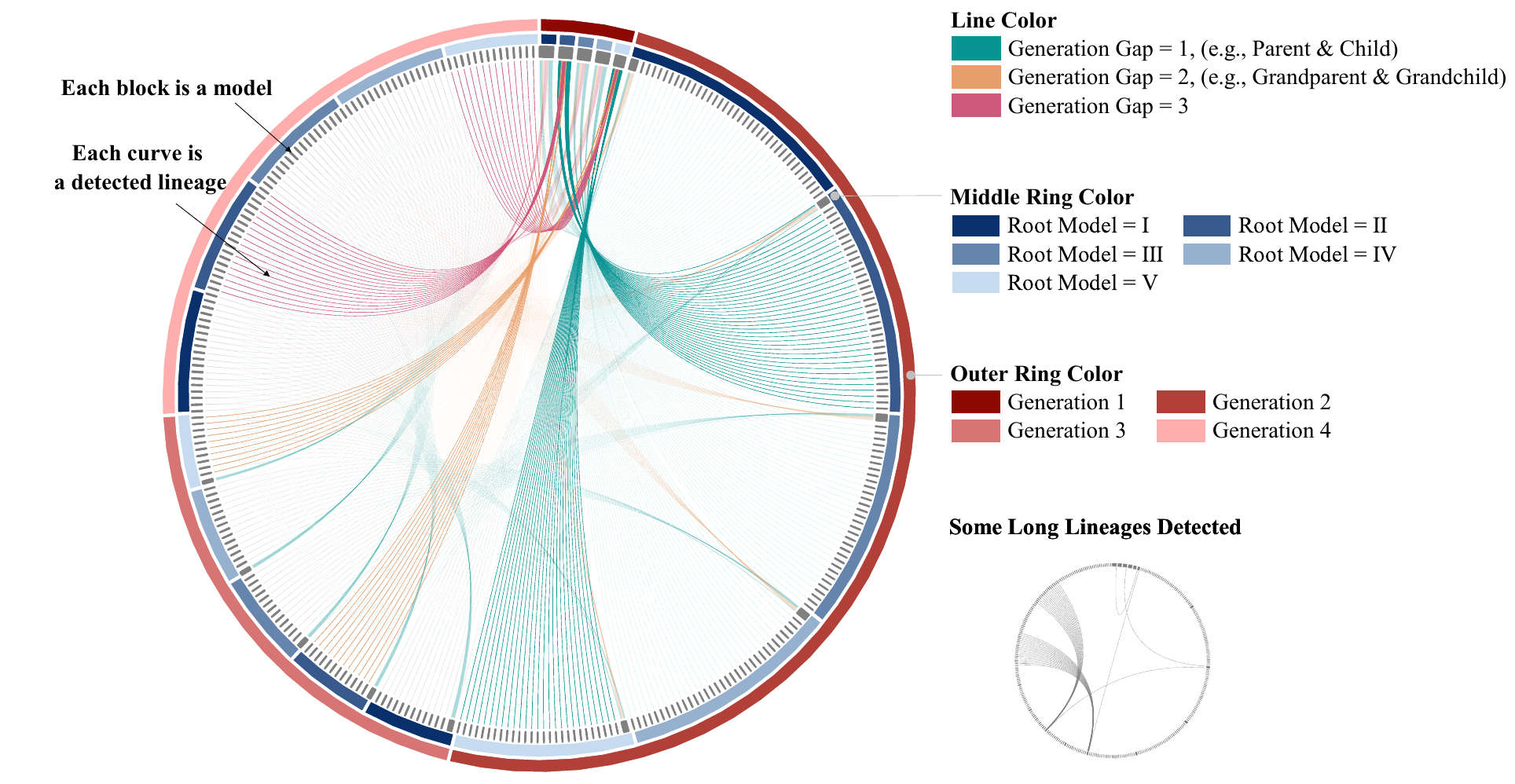}
        \caption{Model-level lineage detection}
        \label{fig:head:task1}
    \end{subfigure}
    \hfill
    \begin{subfigure}[b]{0.4\linewidth}
        \includegraphics[width=\linewidth]{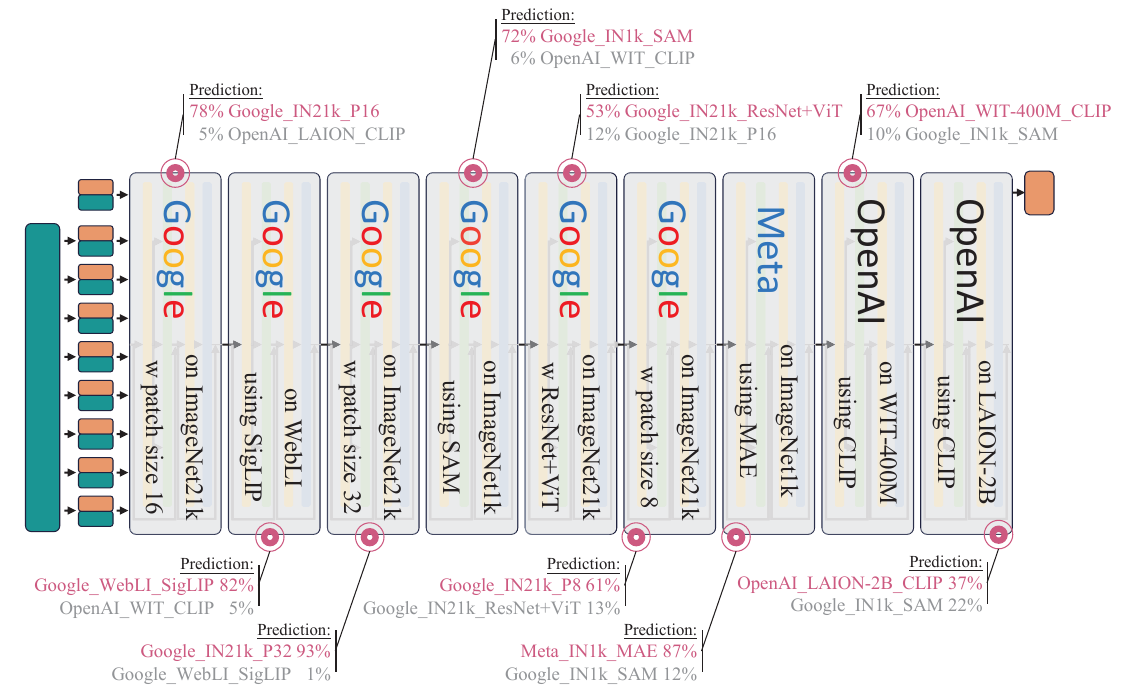}
        \caption{Layer-level lineage detection}
        \label{fig:head:task2}
    \end{subfigure}
    \caption{Neural lineage detection showcase. \cref{fig:head:task} shows the idea of neural lineage detection. \cref{fig:head:task1} presents the detected lineages among 300 networks using the proposed lineage detector. \cref{fig:head:task2} presents the lineage detector's ability to find ancestors of sub-networks. }
    \label{fig:head}
\end{center}%
}]

\begin{abstract}\vspace{-0.5em}
Given a well-behaved neural network, is possible
to identify its parent, based on which it was tuned?
In this paper, we introduce a novel task known as {neural lineage} detection, aiming at discovering lineage relationships between parent and child models. Specifically, from a set of parent models, neural lineage detection predicts which parent model a child model has been fine-tuned from. We propose two approaches to address this task. (1) For practical convenience, we introduce a learning-free approach, which integrates an approximation of the finetuning process into the neural network representation similarity metrics, leading to a similarity-based lineage detection scheme. (2) For the pursuit of accuracy, we introduce a learning-based lineage detector comprising encoders and a transformer detector. Through experimentation, we have validated that our proposed learning-free and learning-based methods outperform the baseline in various learning settings and are adaptable to a variety of visual models. Moreover, they also exhibit the ability to trace cross-generational lineage, identifying not only parent models but also their ancestors. \let\thefootnote\relax\footnotetext{$^{\dagger}$ Corresponding author.}
\end{abstract}
\vspace{-2em}    
\section{Introduction}
\label{sec:intro}
Over the past decade, deep learning technology has undergone rapid evolution, amassing vast research and applications. Today, the development of deep learning exhibits several new phenomena. First, the ``Pre-Training-Finetuning'' framework has supplanted ``Training from Scratch''.
Deep learning models are no longer isolated; downstream-task models have established a significant dependency relationship with pre-trained models. 
Secondly, with the construction of ``Machine Learning as a Service'' (MLaaS), extensive deep learning communities, and vast neural network model repositories have been established. Numerous deep-learning models tailored for various applications have been developed.
These changes indicate that we are now creating a network of deep learning models characterized by a vast number, differentiation, diversity, and strong inter-generational dependencies among models. We are now, 
poised to consider how to study or leverage this burgeoning network of deep learning models.
Foremost among the tasks is to distinguish the generational relationships between models and reveal this underlying network describing the connections between models. 

We define the task of model lineage detection, illustrated in \cref{fig:head:task}, and investigate it in this work.
Specifically, given a model $f_c$, performing model lineage detection for $f_c$ involves identifying a model $f_p$, from which $f_c$ is fine-tuned\footnote{In this work, we always consider tuning all the parameters.}, among a set of candidate models $\{f_p^{(m)}\}_{m=1}^M$. We collectively refer to the candidate models $f_p^{(m)}$'s as parent models, and the model $f_c$ for which we need to perform lineage detection as a child model.

This task holds significant practical importance in the realms of model reuse, intellectual property protection, and model regulation. For the users of a model, neural lineage detection aids in understanding the model's knowledge inherited from parent models, revealing its generalizability and robustness, uncovering its potential fairness and bias issues, and, thus, enabling more targeted adaptation and improvement of the model. For model developers, the generational relationships provided by neural lineage detection can serve as evidence for intellectual property protection. For third-party organizations, neural lineage detection enhances model traceability, becoming a tool for accountability and regulation.

We propose one learning-free and one learning-based method to detect neural lineage. 
The first, the learning-free method, discussed in \cref{sec:method1},  identifies the most probable parent-child model pair through model similarity. While directly measuring the distance between parent and child models using model similarity metrics is feasible, this approach does not leverage one key piece of information: in neural lineage detection, the child model is fine-tuned from the parent model. Instead, based on the Neural Tangent Kernel (NTK) theory, we utilize neural network linearization to approximate the fine-tuning process and identify a proximate child model. We then compare the distance between this approximated child model and the actual one. 
This idea can be applied across various commonly used similarity metrics to establish a set of model similarity metrics embedded with an approximation of the fine-tuning process. Additionally, we address the computational complexity of neural network linearization using the Taylor expansion method, which is summarized in \cref{prop}.

The second, training-based approach, discussed in \cref{sec:method2},  trains a neural network classifier that incorporates multiple encoder blocks to encode the parameters and (or) features of both parent and child networks, and a prediction block based on the transformer architecture to output the probability of the child model being fine-tuned from each parent neural network. Unlike the learning-free method, which explicitly estimates the fine-tuning process, this learning-based approach relies on the neural network's ability to learn the changes that the fine-tuning process imparts on the parameters and features. 

The contributions of this work are:
\begin{itemize}[itemsep=0pt,topsep=0pt,parsep=0pt]
    \item We explore a novel and significant task, namely neural lineage detection, aiming at detecting the parent model from which a child model is fine-tuned.
    \item We propose two methodologies: a learning-free approach, which can be used for any model and at any time, and a learning-based one with higher prediction accuracy.
    \item Experimental results demonstrate the versatility of our method across classification, segmentation, and detection models; in scenarios of few-shot learning and data imbalance; and for cross-generational lineage detection. 
\end{itemize}

\textbf{Illustrative Examples.}
\cref{fig:head:task1} visualizes the lineage detection results for 300 models using the proposed Lineage Detector. Notably, not only all parent-child pairs were accurately identified (plotted with \textcolor{mygreen}{green} curves), but all the cross-generational grandparent-child pairs and great-grandparent-child pairs are also accurately identified (plotted with \textcolor{myorange}{orange} and \textcolor{mypink}{red} curves, respectively). 

In training downstream task models, it's common to assemble multiple pre-trained models and then fine-tune, to amalgamate the functionalities of different models. We constructed a specific case of such ``assembling+finetuning'' strategy to test whether our proposed lineage detector can be applied to sub-network lineage detection. We concatenate transformer layers from different ViT-B models in the timm repository to create a hybrid ViT-B model. This model was then fine-tuned on the CIFAR-10 dataset. Subsequently, the lineage detector was tasked with identifying the origin of each transformer layer. \cref{fig:head:task2} displays the results, where the origins of all transformer layers were accurately identified.
 \section{Related Works}
\subsection{Deep Model IP Protection}
The task most similar to neural lineage detection is deep model IP protection, which enables model owners to identify pirated copies of their protected models. Common methods for deep model IP protection include:
(1) Model watermarking, where a detectable IP identifier is embedded into the neural network through finetuning or retraining
~\cite{wm1,wm2,wm3,wm4,wm5,wm6,wm7,wm8,wm9,wm10,wm11,DeepSigns,TurnWeakness,JingweCVPR23}. 
(2) Model fingerprinting, which involves using gradient optimization or retrieval to find a set of conferrable samples that induce similar behaviors in the source and pirated models, while ensuring distinct behaviors between the source model and irrelevant models
~\cite{IPGuard,fg1,fg2,fg3,fg4,fg5,fg6,fg7}. 

The primary distinctions between neural lineage detection and neural network IP protection are:
\begin{enumerate}[itemsep=0pt,topsep=0pt,parsep=0pt]
    \item IP protection involves a binary classification to determine whether a model is pirated, while neural lineage detection is a multi-class task to identify parent-child pairs.
    \item Solutions for neural network IP protection often depend on external media, like IP identifiers from model watermarking or conferrable samples through model fingerprinting, which entail extra training, optimization, or search processes. In contrast, neural lineage detection seeks to directly identify model lineage without extra encoding, modification, or fingerprint of the model.
    \item IP protection research also considers that model pirates might modify the stolen models, expecting the protection to resist modifications, such as fine-tuning. Conceptually, IP protection methods claim the ability to detect even fine-tuned pirated models. However, the definition of finetuning in the context of IP protection differs significantly from that in neural lineage detection. In neural lineage detection, finetuning is defined as its general practice: given a model for dataset $\mathcal{D}_1$, fine-tuning adapts it to perform the task on dataset $\mathcal{D}_2$, with the requirement that the fine-tuned model performs well on $\mathcal{D}_2$. In contrast, IP protection views fine-tuning as an attack strategy, aiming at circumventing IP protection without compromising the performance on  $\mathcal{D}_1$. This type of fine-tuning attack does not prioritize the performance on $\mathcal{D}_2$, often employing minimal learning rates, few epochs, or a subset of $\mathcal{D}_1$ as $\mathcal{D}_2$,  which differs fundamentally from the general practice of fine-tuning~\cite{DeepSigns,EntangledWatermarks,FrequencyDomainWM,IPGuard,MetaV,rethinkingOwnershipVerif,TriWM,TurnWeakness}. 
\end{enumerate}
\subsection{Neural Network Representation Similarity}
Measuring representation similarity is a fundamental task of deep learning, and numerous methods have been developed. 
Methods based on Canonical Correlation Analysis (CCA) address the high dimensionality of neural network features~\cite{cca1,cca2,XingyiECCV22,XingyiNeurIPS22}, while alignment-based methods resolve inconsistencies in feature dimensions~\cite{align1,align2,align3}. Beyond directly measuring the distance between feature matrices (vectors) or their statistics~\cite{general1,general2}, instance-wise feature similarity can be leveraged to construct similarity matrices~\cite{cka,dc}, graphs~\cite{graph1}, and simplicial complexes~\cite{top1,top2,top3}, or to identify neighbors for each instance in the feature space. Subsequently, the distance between these constructed objects can be measured, or the neighbor information can be utilized for similarity aggregation. 
The similarity of neural networks is also linked to model fine-tuning. Many studies can be viewed as explicitly constraining certain similarity metrics to enhance stability during model fine-tuning or continual learning, thereby reducing the distance between child and parent models~\cite{CotterJGWNYS19,GohCGF16,Fard2016LaunchAI,LiH16}. In our learning-free method, we do not directly employ previous similarity metrics. Instead, we integrate an estimation of the fine-tuning process into previous similarity metrics and use this new series of metrics to achieve neural lineage detection.
\subsection{Neural Network Linearization}
The study of neural network linearization originated with the Neural Tangent Kernel (NTK)~\cite{JacotHG18,Daniely17} and has been theoretically proven that, as its width increases, the network can be effectively approximated by its first-order Taylor expansion with respect to its parameters at initialization~\cite{LeeXSBNSP19}. Numerous empirical studies have also demonstrated that the linear approximation of neural networks can match or even exceed the performance of nonlinear neural networks, particularly in small-sample regions or specially designed problems~\cite{Ortiz-JimenezMF21,AroraD0SWY20}. Neural network linearization has become a viable alternative to training neural networks in fields such as pruning at initialization~\cite{WangZG20,Gebhart2021AUP,fang2023depgraph,ma2023llmpruner}, training-free neural architecture search~\cite{ChenGW21,WangW0022,Mok2022DemystifyingTN}, training time prediction~\cite{ZancatoARBS20}, dataset distillation~\cite{NguyenNXL21}, and deep model ensemble~\cite{HeLT20}, directly approximating the results of gradient optimization. Our learning-free method can also be seen as an application of this idea. However, beyond this, our proposed method further addresses the efficiency issues associated with the use of neural network linearization.
\section{Similarity-Based Detection}
\label{sec:method1}
\subsection{Notation}
Let $\mathcal{D}$ denote the training dataset of the finetuning process and $\mathcal{X}=\{\bm{x}:(\bm{x},\bm{y})\in\mathcal{D}\}$ and $\mathcal{Y}=\{\bm{y}:(\bm{x},\bm{y})\in\mathcal{D}\}$ denote the inputs and the targets, respectively. Let $N$ denote the number of samples in $\mathcal{D}$ and $\bm{x}^{(i)}$ and $\bm{y}^{(i)}$ denote the $i$-th input and target, respectively. Let $f_p:\mathcal{X}\rightarrow\mathcal{Y}$ and $f_c:\mathcal{X}\rightarrow\mathcal{Y}$ denote the parent neural network and the child neural network parameterized by $\bm{\theta}_p$ and $\bm{\theta}_c$, respectively. We use the vectorization form of both the parameters of the neural network and the output of the neural network to facilitate the matrix representation. For example, we write  $\bm{\theta}_p\in\mathbb{R}^{|\bm{\theta}_p|\times 1}$, where $|\bm{\theta}_p|$ is the number of the parameters in $f_p$, $f_p(\bm{x})\in\mathbb{R}^{K\times1}$, where $K$ is the dimension of the targets. 
Given that we are considering the matching problem between the child model and multiple parent models, we use superscripts to index the parent models. Specifically, $f_p^{(m)}$ is used to denote the $m$-th parent model. Let $\Bar{f}_p:\mathcal{X}\rightarrow\mathcal{Y}$ denote the linear approximation of the fine-tuned model, which will be defined in \cref{sec:method1:method}. Let $\langle\cdot,\cdot\rangle$ denote the matrix inner product. Let $\sg(\cdot)$ denote the stop gradient operator. Let $*$ denote the element-wise multiplication.

\subsection{Method}
\label{sec:method1:method}
The learning-free method, conceptually, consists of two steps: (1) approximation, which approximates the child model obtained through the fine-tuning process, and (2) measurement, which compares the distance between the approximated child model and the actual child model. 

In the approximation step, our method uses the strategy of neural network linearization. Previous work has demonstrated that for wide neural networks, this linear approximation provides an excellent and efficient approximation of the dynamics of the original non-linear neural network under gradient descent algorithms~\cite{lm}. Specifically, given a parent model and a child model, the output of the linearized parent model is defined by its first-order Taylor expansion
\begin{equation}
    \Bar{f}_p(\bm{x}) \triangleq f_p(\bm{x}) + \nabla_{\bm{\theta}_p}f_p(\bm{x})(\bm{\theta}_c - \bm{\theta}_p), \label{def:lm}
\end{equation}
where $f_p(\bm{x})$ denotes the output of the parent model; $\nabla_{\bm{\theta}_p}f_p(\bm{x})$ yields the Jacobian matrix of $f_p(\bm{x})$ with respect to its parameter $\bm{\theta}_p$
; and $\bm{\theta}_c - \bm{\theta}_p$ signifies the parameter change between the child and parent models. If the child model is indeed fine-tuned from the parent model, then the approximated $\Bar{f}_p(\bm{x})$ should closely resemble the output of the child model itself, $f_c(x)$
. If the child model is not fine-tuned from the parent model, the gradient and parameter change in equation \cref{def:lm} would be entirely independent.

\begin{figure}[t]
    \centering
    \begin{subfigure}[b]{0.49\linewidth}
        \includegraphics[width=\linewidth]{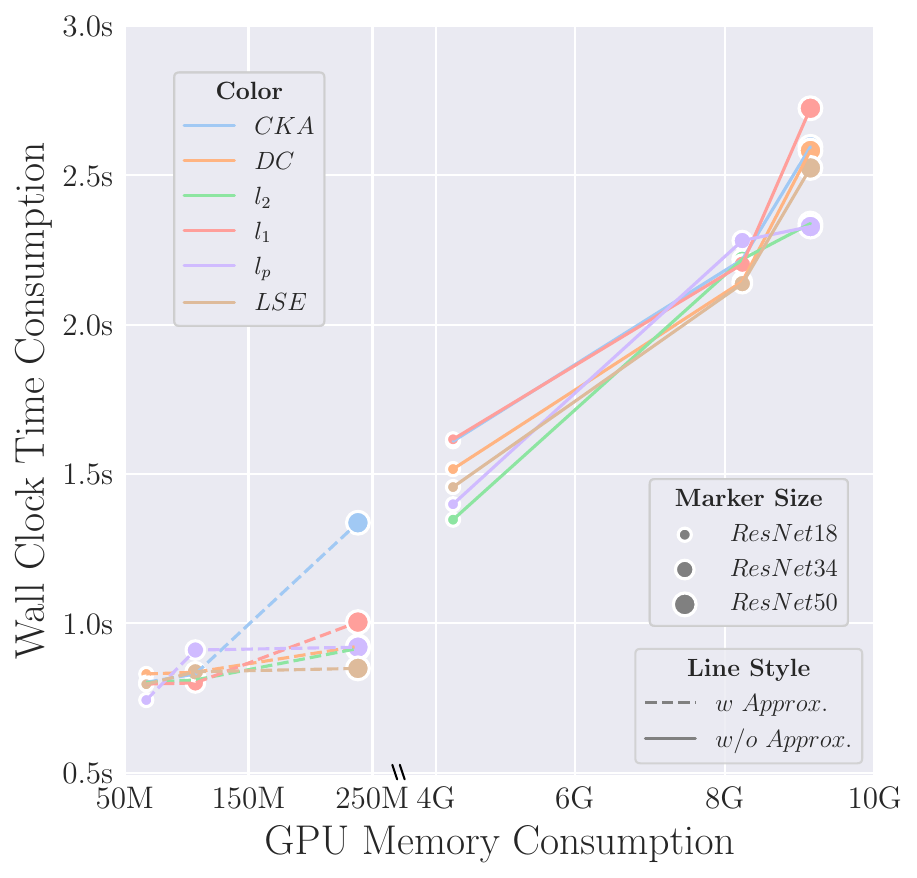}
        \caption{Different model size}
        \label{fig:1:model}
    \end{subfigure}
    \hfill
    \begin{subfigure}[b]{0.49\linewidth}
        \includegraphics[width=\linewidth]{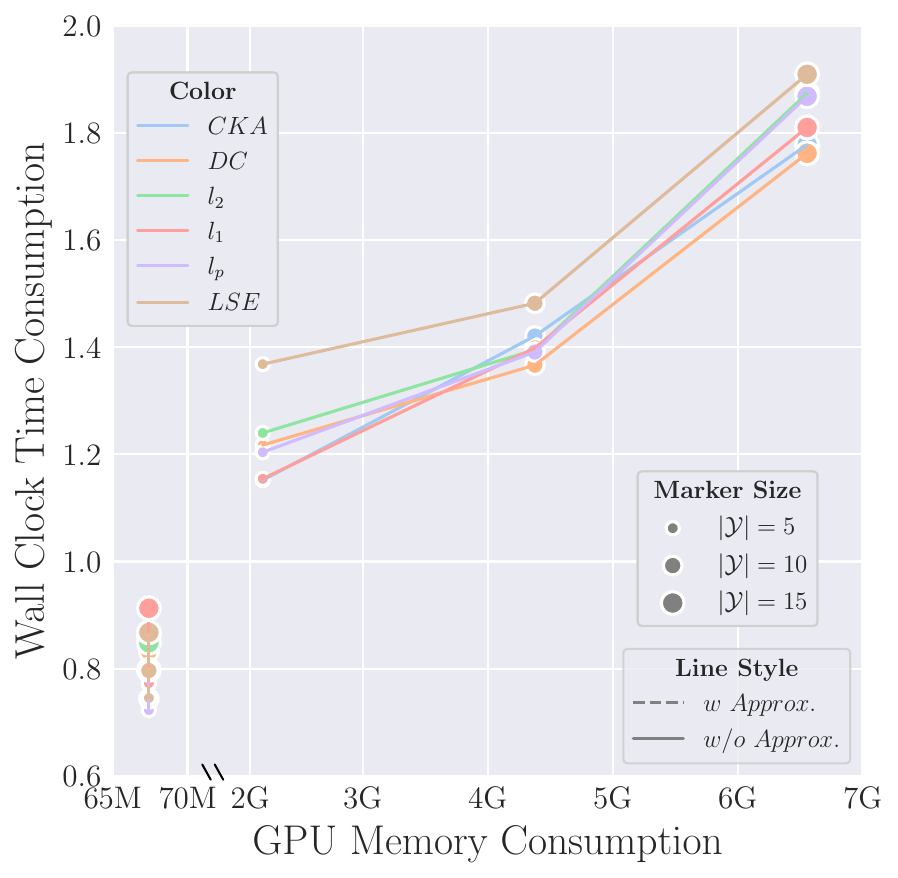}
        \caption{Different output size}
        \label{fig:1:feature}
    \end{subfigure}
    \vspace{-1em}
    \caption{Comparison of the execution time and GPU memory Consumption for methods with and without approximation.}
    \label{fig:1}
    \vspace{-1em}
\end{figure}

In the measurement step, we compute the similarity between $\Bar{f}_p(\bm{x})$ and $f_c(\bm{x})$. However, to obtain $\Bar{f}_p(\bm{x})$, one must first compute the Jacobian matrix. Moreover, most neural network similarity metrics require multiple samples. Thus, for calculating a scalar distance metric, first computing $\Bar{f}_p(\bm{x})$ and then measuring the distance necessitates gradient computations through the entire neural network $N \times K$ times, and the storage of gradient value amounts to $N \times K \times |\bm{\theta}|$, where $N$ is the number of samples and $|\bm{\theta}|$ is the number of parameters in $f_p$. To reduce the computational time and space complexity, instead of this step-by-step approach, we consider integrating the first approximation step with the second distance measurement step with a linear approximation of the similarity metric. We apply this idea to several commonly used similarity metrics and list the approximation results below. Detailed derivations are left in the appendix.

\begin{proposition}[Similarity Approximation.] Define $\bm{d}_i\triangleq \allowbreak f_p(\bm{x}^{(i)})-f_c(\bm{x}^{(i)})$ to be the difference between the outputs of $f_p$ and $f_c$ for input $\bm{x}^{(i)}$.
Let $s(\Bar{f}_p, f_c)$ be the similarity between $\{\Bar{f}_p(\bm{x}^{(i)})\}_{i=1}^N$ and $\{f_c(\bm{x}^{(i)})\}_{i=1}^N$, whose linear approximation has the form of 
\begin{equation}
    s(\Bar{f}_p,\!f_c)\!\approx\!s(f_p,\! f_c)\!+\!\nabla_{\bm{\theta}_p}[\sum_{i=1}^N\sg(\Pi_i)f_p(\bm{x}^{(i)})](\bm{\theta}_c\text{-}\,\bm{\theta}_p)
\end{equation}\label{prop}
\begin{subproposition}[$l_1$ Similarity.]\label{prop:l1}
    For $l_1$ similarity $s(\Bar{f}_p, f_c)\triangleq-\frac{1}{NK}\allowbreak\sum_{i=1}^N||\Bar{f}_p(\bm{x}^{(i)})\!- f_c(\bm{x}^{(i)})||_1$, $\Pi_i=-\frac{1}{NK}\sign(d_i)^T$;
\end{subproposition}

\begin{subproposition}[$l_2$ Similarity.]\label{prop:l2}
    For $l_2$ similarity $s(\Bar{f}_p, f_c)\triangleq-\frac{1}{NK}\allowbreak\sum_{i=1}^N||\Bar{f}_p(\bm{x}^{(i)}) - f_c(\bm{x}^{(i)})||_2^2$, $\Pi_i=
        -\frac{2}{NK}\bm{d}_i^T$;
\end{subproposition}

\begin{subproposition}[$l_p$ Similarity.]\label{prop:lp}
    For $l_p$ similarity $s(\Bar{f}_p, f_c)\triangleq-\frac{1}{NK}\allowbreak\sum_{i=1}^N||\Bar{f}_p(\bm{x}^{(i)}) - f_c(\bm{x}^{(i)})||_p^p$, 
    \begin{equation}
        \Pi_i=
        -\frac{p}{NK}[\sign(\bm{d}_i)*|\bm{d}_i|^{p-1}]^T;
    \end{equation}
\end{subproposition}

\begin{subproposition}[$log$-$sum$-$exp$ Similarity.]\label{prop:log-sum-exp}
    For LSE similarity $s(\Bar{f}_p, f_c)\triangleq-\frac{1}{NKt}\sum_{i=1}^N\log\sum_{k=1}^Ke^{t|\Bar{f}_p(\bm{x}^{(i)})\!-\!f_c(\bm{x}^{(i)})|}$, 
    \begin{equation}
        \Pi_i=
        -\frac{1}{NK}[\sign(\bm{d}_i)*\softmax(t|\bm{d}_i|)]^T;
    \end{equation}
\end{subproposition}

\begin{subproposition}[CKA.]\label{prop:cka}
    For squared Centered Kernel Alignment with linear kernel defined in ~\cite{cka}, 
    $\Pi_i=s(f_p, f_c)\zeta_i$;
\end{subproposition}

\begin{subproposition}[DC.]\label{prop:dc}
    For squared Distance Correlation defined in~\cite{dc}, 
    $\Pi_i=s(f_p, f_c)\xi_i$;
\end{subproposition}
where the complete expression of $\zeta_i$ and $\xi_i$ are presented in the Appendix.
\end{proposition}

\begin{figure}[t]
    \centering
    \begin{subfigure}[b]{0.49\linewidth}
        \includegraphics[width = \textwidth]{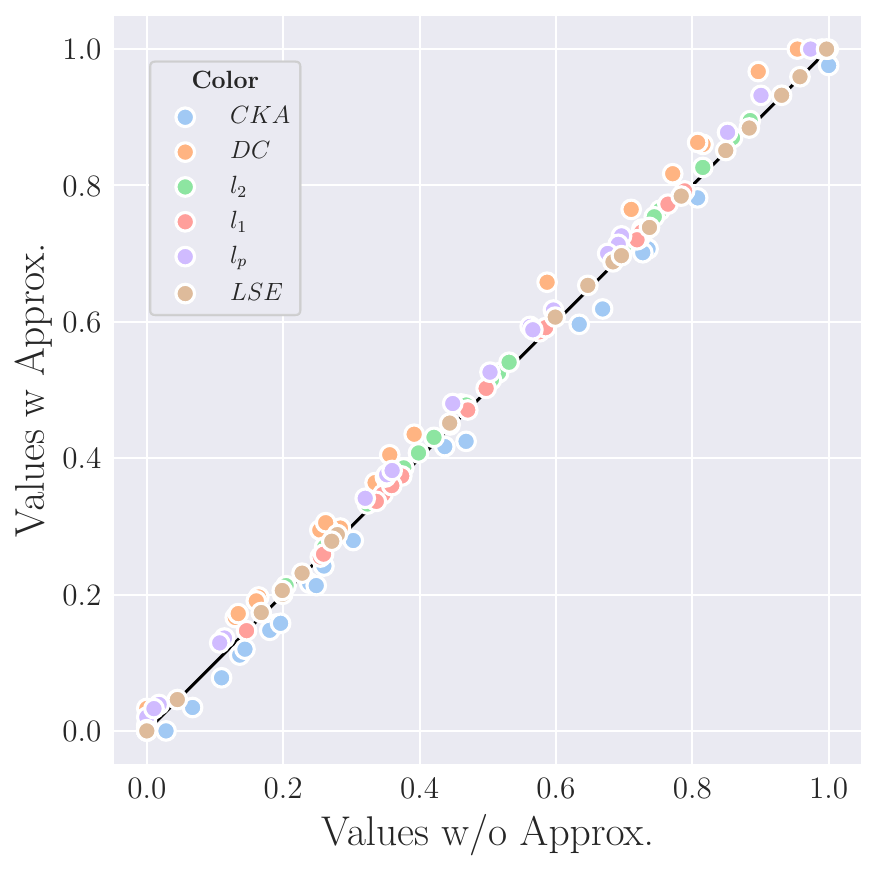}
        \caption{Similarity values comparison}
        \label{fig:2:value}
    \end{subfigure}
    \hfill
    \begin{subfigure}[b]{0.49\linewidth}
        \includegraphics[width = \textwidth]{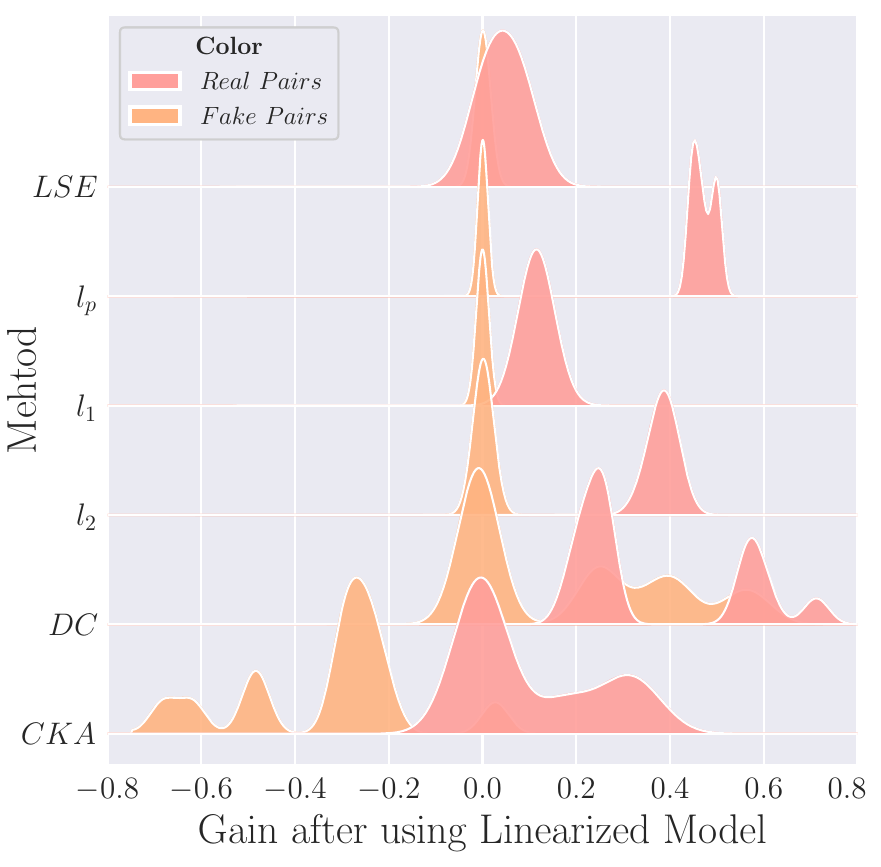}
        \caption{Similarity gain density}
        \label{fig:2:gain}
    \end{subfigure}\vspace{-1em}
    \caption{The influence of approximations and linearized model. \cref{fig:2:value} plots the similarity values measured with and without using approximation. \cref{fig:2:gain} plots the similarity gains after using the linearized model to get a better anchor.}
    \label{fig:2}\vspace{-1em}
\end{figure}

\begin{remark}
    The approximations in \cref{prop} reduce the computational complexity.
    Computing the approximated similarity metrics necessitates only a single back-propagation through the neural network, and each neural network parameter requires the storage of just one gradient value. 
    \cref{fig:1} compares the wall-clock time and GPU memory consumption for computing model similarity using our method versus a step-by-step approach. Our method significantly reduces GPU memory consumption and also requires less computation time. Notably, since our approach first computes a weighted sum of model outputs before evaluating the derivative, GPU memory consumption does not increase with the size of the model's output. 
\end{remark}
\begin{remark}
    In \cref{prop}, the gradient calculations in the previous step-by-step method have been consolidated. All approximated similarity metrics solely encompass one gradient computation of one scalar. 
    This scalar is a weighted sum of the neural network's output, where different similarity metrics employ distinct weighting schemes. For instance, the $\pi_i$ of $l_2$ similarity is the difference in output between the parent and child models, suggesting that $l_2$ similarity
    focuses more on features with significant discrepancies between the parent and child models, while down-weighting features that are similar between them. A similar observation can be made for the approximation of the $log$-$sum$-$exp$ similarity, where $\Pi_i$ is the softmax score of the difference in output between the parent and child models.
\end{remark}
\begin{remark}
    Given that $l_\infty$ is inherently non-differentiable, it precludes its Taylor expansion. 
    The $p$-norm and $log$-$sum$-$exp$ in \cref{prop} are approximations to the $l_{\infty}$ similarity.
\end{remark}
\begin{remark}
    We also empirically assessed the accuracy of the approximation in the proposition, with relevant results displayed in \cref{fig:2:value}. The x-axis of Fig. 2(a) represents the similarity values computed using the step-by-step approach, while the y-axis represents the similarity values obtained using the method in \cref{prop}. All data points are closely located around the identity line, indicating that the error introduced by the Taylor expansion in the proposition is minimal. Therefore, considering the reduction in computational complexity brought about by the approximation, it offers a more advantageous approach overall compared to the step-by-step method.
\end{remark}

Thus far, given a child model $f_c$ for lineage detection and a set of parent models $\{f_p^{(i)}\}_{i=1}^M$, we have the similarity metrics $\{s(\bar{f}_p,f_c)\}_{m=1}^M$ based on the linear approximation of the fine-tuning.
From it, a matching probability vector $P\in[0,1]^M$ can be calculated as the softmax of the similarity measures with $P_m\triangleq\frac{e^{s(\bar{f}_p^{(m)},f_c)}}{\sum_{l=1}^Me^{s(\bar{f}_p^{(l)},f_c)}}$. Finally, the prediction can be drawn by selecting the parent model with the highest matching probability. 

To enhance the adaptability across diverse models, we introduce two adjustments in our approach. First, recognizing that not every model aligns seamlessly with the theoretical underpinnings of NTK, we substitute the fixed \cref{def:lm} with $\Bar{f}_p(\bm{x}) \triangleq f_p(\bm{x}) + \alpha\nabla{\bm{\theta}_p}f_p(\bm{x})(\bm{\theta}_c - \bm{\theta}_p)$, where $\alpha$ is a hyper-parameter, analogous to the learning rate. Second, we extend the approximation and similarity measurement to the intermediate layer features, rather than confining them solely to the neural network's final output.

\subsection{Intuitive Explanation}
\begin{proposition}[The Optimality of Measurement]\label{prop:optapp} Let $f'_p(\bm{x}) \triangleq f_p(\bm{x}) + \bm{W}(\bm{x})(\bm{\theta}_c - \bm{\theta}_p)$ and $f''_p(\bm{x}) \triangleq f_p(\bm{x}) + \nabla_{\bm{\theta}_p}f_p(\bm{x})\bm{Z}$ denote two linearized approximation of parent model with undetermined parameters $\bm{W}(\bm{x})\in\mathbb{R}^{K\times|\bm{\theta}|}$ for each $\bm{x}$ and $\bm{Z}\in\mathbb{R}^{|\bm{\theta}|\times 1}$ for all $\bm{x}$, respectively.
For the real parent-child model pair $(f_p, f_c)$ and $l_2$ similarity,
\begin{align}
    s(\Bar{f}_p,f_c) = \max_{\bm{W}(\bm{x}),\bm{x}\in\mathcal{X}} s(f_p',f_c)= \max_{\bm{Z}} s(f_p'',f_c)
\end{align}
\end{proposition}
This proposition demonstrates that, when allowing for a linear approximation of the finetuning process, under the \(l_2\) similarity metric, our method yields the optimal results, maximizing the similarity between the real parent-child pair. 
To elaborate further, the first optimization problem suggests that for each $\bm{x}$, our method implicitly identifies a hyperplane passing through $f_p(\bm{x})$ such that the projection of $f_c(\bm{x})$ onto this hyperplane minimizes the distance to the point $(\bm{\theta}_c - \bm{\theta}_p)$ on the hyperplane; the second optimization problem indicates that our method first implicitly determines the projection of $\{f_c(\bm{x}^{(i)})\}_{i=1}^N$ onto the hyperplane constructed by $\{\nabla_{\bm{\theta}_p}f_p(\bm{x}^{(i)})\}_{i=1}^N$, and then measures the distances between $\{f_c(\bm{x}^{(i)})\}_{i=1}^N$ and its projection.

Given that the aforementioned proposition only theoretically analyzed how linearization in \cref{def:lm} impacts the similarity between real parent-child pairs, we further empirically verify if the intuition derived from the proposition holds true, and provide insights into how linearization affects the similarity between fake parent-child pairs. \cref{fig:2:gain} illustrates the changes in similarity after linearization. For real parent-child pairs, the similarity metric generally increases, which is consistent with our theoretical findings. For fake parent-child pairs, the linearized model causes two kinds of phenomenon: it either generally reduces the similarity or has minimal impact on the similarity metric. The reasons for these phenomena might be that the linearized model essentially predicts the fine-tuning process starting from the parent model based on the gradient direction and uses the predicted new model as an anchor to compare its similarity with the child model. If the angle between the gradient generated by the fake parent model and the actual fine-tuning direction is obtuse, the prediction of fine-tuning results in an anchor even farther from the child model than the previous fake parent, leading to a decrease in similarity. If the gradient generated by the fake parent model is orthogonal to the actual fine-tuning direction, the distance from the new anchor to the child model remains almost the same as the distance from the previous parent model to the child model, resulting in an almost unchanged similarity metric.
\section{Lineage Detector}
\label{sec:method2}
\begin{figure}[t]
    \centering
    \includegraphics[width=\linewidth]{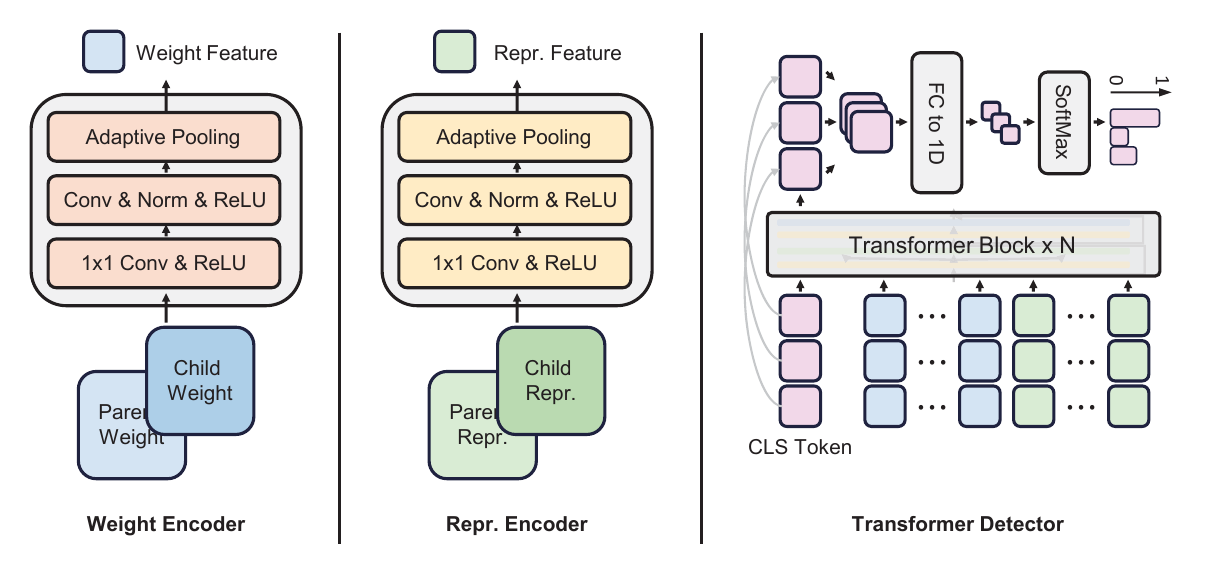}\vspace{-1em}
    \caption{The architecture of the proposed lineage detector. Weights and features are first encoded separately and are then fed into a transformer to obtain the prediction score.}
    \label{fig:detector}\vspace{-1em}
\end{figure}
In this section, we introduce the learning-based lineage detection method. The proposed lineage detector can be viewed as a nonlinear improvement of the method described in \cref{sec:method1}. Here, the effects of finetuning are directly learned by the lineage detector, rather than being approximated by a fixed linear relationship. The lineage detector, shown in \cref{fig:detector}, comprises a weight encoder, a feature encoder, and a transformer detector, which takes the weights and features of parent and child models as input to predict their lineage probability. The workflow of the lineage detector is detailed as follows. 

First, parameters $\bm{\theta}_p$ and features $\bm{F}_p$ from a parent model are first vectorized and then reshaped into matrices, such that, $\bm{\theta}_p\in\mathcal{R}^{H_{\theta}\times W_{\theta}}$ and $\bm{F}_p\in\mathcal{R}^{H_{F}\times W_{F}}$. The widths and heights are chosen based on the architecture of the parent model. The weights and features of a child model are also pre-processed to have the same shapes, so that, $\bm{\theta}_c\in\mathcal{R}^{H_{\theta}\times W_{\theta}}$ and $\bm{F}_c\in\mathcal{R}^{H_{F}\times W_{F}}$. The weights from the parent and child are then stacked together as input for the weight encoder, while their features are similarly stacked as the input of the feature encoder. We define the stacked weights and features as $\bm{\theta}\triangleq [\bm{\theta}_p;\bm{\theta}_c]\in\mathcal{R}^{2\times H_{\theta}\times W_{\theta}}$ and $\bm{F}\triangleq [\bm{F}_p;\bm{F}_c]\in\mathcal{R}^{2\times H_{F}\times W_{F}}$, respectively. Compared to encoding the parent's weights(features) and the child's weight(feature) independently, this stacking strategy facilitates the model's direct learning of the element-level relationships between parent and child models, since the parent's and child's weights(features) are naturally aligned and the finetuning process can bring subtle changes in the weights(features) that can be missed if encoded separately. 

The weight encoder $\Phi_{\bm{\theta}}$ and the feature encoder $\Phi_{\bm{F}}$ share the same CNN architecture. In the first layer, 1x1 kernels are used to further facilitate the model's extraction of the element-level relationships. The intermediate layers of the convolutional network are standard convolutions combined with batch normalization and ReLU activation. After the convolutional layers, adaptive average pooling, with output size $1\times 1$, is used to address the potential inconsistencies in the shape of the $\bm{\theta}$ and $\bm{F}$. Squeezing the redundant dimension, the outputs of the encoders are vectors, $z_{\bm{\theta}}\triangleq \Phi_{\bm{\theta}}(\bm{\theta})\in\mathcal{R}^{D}$ and $z_{\bm{F}}\triangleq \Phi_{\bm{F}}(\bm{F})\in\mathcal{R}^{D}$, where $D$ is the number of the output channels of the CNN and the token size of the following transformer $\Psi$.

Similar to the position embedding used in transformers, we add learnable weight embedding $E_{\bm{\theta}}\in\mathcal{R}^{D}$ and feature embedding $E_{\bm{\theta}}\in\mathcal{R}^{D}$ to distinguish weight and feature for subsequent processing by the transformer. These are then concatenated with a learnable classification (cls) token $z_{cls}\in\mathcal{R}^{D}$ and fed into a multi-layer transformer. 
A transformer is used because the weights and features change interdependently during the finetuning process, and the attention mechanism allows the final prediction to consider both the information from weights and features and their interactive effects. Furthermore, the transformer structure is adaptable to scenarios with multiple weight tokens, multiple feature tokens, or only one of the weight and feature tokens. Finally, a linear head transforms the final output at the position of the class token into the final score, which can be written as
\begin{align}
    z^{final} &= \Psi([z_{cls};z_{\theta}+E_{\bm{\theta}};z_{F}+E_{\bm{F}}]),\\
    s&=Head(z_{cls}^{final}).
\end{align}

For a set of parent models $\{f_p^{(m)}\}_{m=1}^M$, a set of score $\{s^{(m)}\}_{m=1}^M$ can be obtained for each parent-child pair, and the model lineage probability prediction can be derived from the softmax of these scores. Given the true parent index, the lineage detector can be trained using a cross-entropy loss.
\begin{table*}[]
\small
\resizebox{1\textwidth}{!}{
\begin{tabular}{@{}cc@{$\ $}c@{$\quad$}r@{$\pm$}l@{$\ $}r@{$\pm$}l@{$\ $}r@{$\pm$}l@{$\ $}r@{$\pm$}l@{$\ $}r@{$\pm$}l@{$\ $}r@{$\pm$}l@{$\ $}r@{$\pm$}l@{$\ $}r@{$\pm$}l@{$\ $}r@{$\pm$}l@{}}
\toprule
                                          &                                        &                 & \multicolumn{2}{c}{FMNIST} & \multicolumn{2}{c}{EMNIST} & \multicolumn{2}{c}{Cifar10} & \multicolumn{2}{c}{SVHN} & \multicolumn{2}{c}{DTD} & \multicolumn{2}{c}{Pet} & \multicolumn{2}{c}{\makecell{Cifar10\\5shot}} & \multicolumn{2}{c}{\makecell{Cifar10\\50shot}} & \multicolumn{2}{c}{\makecell{Cifar10\\\scriptsize{Imbalanced}}} \\ \midrule
\parbox[t]{2mm}{\multirow{12}{*}{\rotatebox[origin=c]{90}{Fully Connected Network}}} & \parbox[t]{2mm}{\multirow{5}{*}{\rotatebox[origin=c]{90}{w/o Approx.}}} & $l_1$           & 24.11        & 1.27        & 35.97        & 1.25        & 70.04         & 1.05        & 84.09       & 1.64       & \multicolumn{2}{c}{-}   & \multicolumn{2}{c}{-}   & 80.31            & 1.22           & 73.93            & 1.01            & 75.42              & 1.43              \\
                                          &                                        & $l_2$           & 23.98        & 2.04        & 45.61        & 1.59        & 76.88         & 0.64        & 68.84       & 1.71       & \multicolumn{2}{c}{-}   & \multicolumn{2}{c}{-}   & 86.37            & 1.02           & 75.32            & 1.18            & 78.24              & 1.72              \\
                                          &                                        & $l_{\infty}$ & 11.84        & 1.23        & 37.99        & 1.83        & 77.32         & 0.81        & 70.06       & 2.25       & \multicolumn{2}{c}{-}   & \multicolumn{2}{c}{-}   & 78.79            & 1.44           & 75.14            & 1.08            & 76.41              & 1.24              \\
                                          &                                        & $CKA$           & 4.39         & 0.37        & 6.58         & 0.76        & 52.49         & 1.31        & 29.92       & 1.27       & \multicolumn{2}{c}{-}   & \multicolumn{2}{c}{-}   & 18.18            & 1.83           & 21.21            & 0.97            & 50.76              & 1.72              \\
                                          &                                        & $DC$            & 3.51         & 0.49        & 6.32         & 0.91        & 54.92         & 1.22        & 29.92       & 0.87       & \multicolumn{2}{c}{-}   & \multicolumn{2}{c}{-}   & 19.71            & 1.71           & 21.81            & 1.52            & 50.38              & 1.61              \\ \cmidrule(l){2-21} 
                                          & \parbox[t]{2mm}{\multirow{6}{*}{\rotatebox[origin=c]{90}{w Approx.}}}    & $l_1$           & 25.44        & 1.27        & 36.40         & 1.31        & 71.59         & 1.12        & \underline{85.22}       & \underline{1.57}       & \multicolumn{2}{c}{-}   & \multicolumn{2}{c}{-}   & 85.61            & 2.51           & 74.62            & 1.12            & 75.76              & 1.51              \\
                                          &                                        & $l_2$           & \underline{26.75}        & \underline{1.75}        & \underline{46.93}        & \underline{1.73}        & \underline{80.03}         & \underline{0.71}        & 70.07       & 1.71       & \multicolumn{2}{c}{-}   & \multicolumn{2}{c}{-}   & \underline{88.64}            & \underline{1.31}           & 75.38            & 1.18            & \underline{78.79}              & \underline{1.72}              \\
                                          &                                        & $l_p$           & 21.05        & 1.69        & 41.23        & 1.79        & 78.79         & 0.83        & 70.83       & 1.61       & \multicolumn{2}{c}{-}   & \multicolumn{2}{c}{-}   & 86.73            & 1.41           & 69.71            & 1.51            & 76.51              & 1.16              \\
                                          &                                        & $LSE$           & 26.32        & 1.27        & 37.28        & 1.36        & 73.48         & 0.78        & 71.59       & 1.79       & \multicolumn{2}{c}{-}   & \multicolumn{2}{c}{-}   & 85.61            & 1.39           & \underline{75.76}            & \underline{1.05}            & 78.03              & 1.47              \\
                                          &                                        & $CKA$           & 5.71         & 0.48        & 14.91        & 1.04        & 53.03         & 1.16        & 57.95       & 0.98       & \multicolumn{2}{c}{-}   & \multicolumn{2}{c}{-}   & 21.21            & 2.41           & 21.97            & 0.96            & 51.14              & 1.69              \\
                                          &                                        & $DC$            & 14.48        & 1.41        & 7.64         & 0.81        & 61.74         & 0.91        & 28.41       & 1.01       & \multicolumn{2}{c}{-}   & \multicolumn{2}{c}{-}   & 25.01            & 2.42           & 36.74            & 1.89            & 51.14              & 1.41              \\ \cmidrule(l){2-21} 
                                          & \multicolumn{2}{c}{Lineage Detector}                     & \textbf{97.86}        & \textbf{1.19}\mystar        & \textbf{93.61}        & \textbf{2.37}\mystar        & \textbf{99.11}         & \textbf{0.36}\mystar        & \textbf{99.31}       & \textbf{0.35}\mystar       & \multicolumn{2}{c}{-}   & \multicolumn{2}{c}{-}   & \textbf{99.94}            & \textbf{0.35}\mystar           & \textbf{98.84}            & \textbf{1.43}\mystar            & \textbf{99.35}              & \textbf{0.24}\mystar              \\ \midrule
\parbox[t]{2mm}{\multirow{12}{*}{\rotatebox[origin=c]{90}{ResNet18}}}                & \parbox[t]{2mm}{\multirow{5}{*}{\rotatebox[origin=c]{90}{w/o Approx.}}} & $l_1$           & 90.37        & 0.82        & 90.47        & 0.84        & 90.19         & 1.33        & 98.75       & 0.72       & 89.41       & 1.66      & 90.23       & 0.64      & 98.85            & 0.19           & 98.68            & 0.88            & 87.50              & 1.11              \\
                                          &                                        & $l_2$           & 86.84        & 1.15        & 87.51        & 1.36        & 90.10         & 1.41        & 98.64       & 0.22       & 88.68       & 1.66      & 89.43       & 0.99      & 97.75            & 0.19           & 98.41            & 0.41            & 87.68              & 0.11              \\
                                          &                                        & $l_{\infty}$ & 78.57        & 0.87        & 95.83        & 0.65        & 79.76         & 0.64        & 95.24       & 0.41       & 88.24       & 2.01      & 85.81       & 1.45      & 98.07            & 0.69           & 98.93            & 0.51            & 79.17              & 1.21              \\
                                          &                                        & $CKA$           & 68.45        & 2.49        & 75.59        & 2.19        & 45.83         & 0.81        & 63.54       & 1.93       & 62.94       & 1.52      & 66.04       & 1.98      & 71.11            & 3.45           & 64.82            & 0.52            & 62.51              & 1.48              \\
                                          &                                        & $DC$            & 58.92        & 2.21        & 32.14        & 1.61        & 61.33         & 2.71        & 45.43       & 1.68       & 62.94       & 1.75      & 51.47       & 3.17      & 67.78            & 2.28           & 65.56            & 0.51            & 60.47              & 1.59              \\ \cmidrule(l){2-21} 
                                          & \parbox[t]{2mm}{\multirow{6}{*}{\rotatebox[origin=c]{90}{w Approx.}}}    & $l_1$           & \underline{98.81}        & \underline{0.41}        & \underline{98.92}        & \underline{0.59}        & \underline{94.64}         & \underline{0.63}        & 98.81       & 0.41       & \underline{94.71}       & \underline{0.32}      & \underline{95.27}       & \underline{0.84}      & \underline{99.89}            & \underline{0.62}           & 98.93            & 0.61            & \underline{95.84}              & \underline{0.99}              \\
                                          &                                        & $l_2$           & 88.69        & 0.65        & 88.09        & 1.08        & 90.47         & 0.84        & 97.94       & 1.26       & 90.59       & 1.75      & 90.53       & 0.63      & 98.89            & 0.59           & \underline{99.36}            & \underline{0.52}            & 88.09              & 0.93              \\
                                          &                                        & $l_p$           & 80.95        & 0.48        & 80.95        & 1.16        & 81.55         & 0.68        & 94.05       & 0.92       & 91.18       & 1.87      & 89.66       & 0.35      & 92.22            & 1.86           & 93.54            & 2.23            & 80.95              & 0.85              \\
                                          &                                        & $LSE$           & 96.43        & 0.97        & 98.21        & 0.41        & 89.29         & 1.21        & \underline{99.41}       & \underline{0.33}       & 13.53       & 1.43      & 11.24       & 1.21      & 98.82            & 0.22           & 97.85            & 1.21            & 91.67              & 1.23              \\
                                          &                                        & $CKA$           & 70.23        & 2.54        & 77.38        & 1.93        & 47.62         & 0.94        & 64.88       & 1.89       & 66.47       & 0.83      & 68.05       & 1.61      & 72.22            & 3.25           & 68.81            & 3.71            & 66.08              & 2.42              \\
                                          &                                        & $DC$            & 60.11        & 2.33        & 34.52        & 0.85        & 63.09         & 2.73        & 45.83       & 1.47       & 63.12       & 1.43      & 52.07       & 2.73      & 67.89            & 2.18           & 65.58            & 2.64            & 60.71              & 1.09              \\ \cmidrule(l){2-21} 
                                          & \multicolumn{2}{l}{Lineage Detector}                     & \textbf{99.21}        & \textbf{0.16}        & \textbf{99.32}        & \textbf{0.05}        & \textbf{99.87}         & \textbf{0.07}\mystar        & \textbf{99.64}       & \textbf{0.12}       & \textbf{99.59}       & \textbf{0.47}\mystar      & \textbf{99.74}       & \textbf{0.21}\mystar      & \textbf{99.91}            & \textbf{0.52}           & \textbf{99.38}            & \textbf{0.02}            & \textbf{97.01}              & \textbf{0.41}              \\ \bottomrule
\end{tabular}
}
\vspace{-1em}
  \caption{Lineage detection results for classification tasks. The best (the second-best) ones are marked in \textbf{bold} (\underline{underlined}). * indicates that the lineage detector has a statistically significant improvement compared to the best among other methods, with a p-value less than 0.05.}
  \label{tab:main}\vspace{-1.5em}
\end{table*}
\section{Experiments}
In this section, we present our core experimental results, with more results and implementation details reported in the appendix.

\subsection{Classification}
\textbf{Basic setup.} The basic setup focuses on classification. We examined two architectures: a 3-layer fully connected(FC) network with ReLU activation and no normalization, and ResNet18~\cite{resnet}. The simple FC network was included in the experiments as it better aligns with the theoretical hypotheses of Neural Network Linearization, allowing us to observe the performance of the learning-free method under relatively ideal conditions. For the FC network, 20 networks trained from scratch on MNIST~\cite{mnist} or 12 on CIFAR100~\cite{cifar} were used as parent models. For ResNet, 7 networks pre-trained on ImageNet~\cite{imagenet} in timm~\cite{timm} package were used. Child models were generated by finetuning parent models on downstream datasets with varying hyperparameters and random seeds. The downstream datasets included FMNIST~\cite{fmnist}, EMNIST-Letters~\cite{emnist}, CIFAR10~\cite{cifar}, Pet~\cite{pet}, and DTD~\cite{dtd}. The average number of child models is 204 per dataset-network structure pair. We compared the accuracy of lineage detection of the proposed learning-free and learning-based methods against a baseline that directly uses parent-child model similarity.
For training the lineage detector, child models were randomly split into training, validation, and test sets in a 7:1:2 ratio, and the performance on the test set was recorded and averaged over 5 repetitions. The results for both learning-free and learning-based methods are averages from 5-fold experiments.

The results are shown in \cref{tab:main}. Key observations include: (1) Generally, the learning-based method significantly outperforms the others with an average accuracy improvement of 36.26\%, and the learning-free method generally outperforms the baseline by an average of 3.17\%, with a maximum improvement of 28\%. (2) The performance of the learning-free method depends on the similarity metric used. No single metric consistently outperforms others, but generally, norm distances perform better than advanced neural network similarity metrics, possibly because these neural network similarity metrics filter out differences that can be eliminated through projection or rotation~\cite{cka,dc}, which are often key clues for lineage detection.

\textbf{Few-shot and imbalanced data.} We also tested our methods on CIFAR10 -5-shot, -50-shot, and -imbalanced datasets\footnote{constructed by down-sampling the last 5 classes of CIFAR10 to 10\%}. The scale of these experiments was comparable to the previous CIFAR10 experiments. Overall, both learning-free and learning-based methods performed better in few-shot scenarios than on complete datasets, likely due to the quicker convergence and fewer samples, making the finetuning process easier to approximate or learn. The advantages of the proposed methods persisted on the few-shot and imbalanced datasets.

\begin{figure}[t]
    \centering
    \begin{subfigure}[b]{0.49\linewidth}
        \includegraphics[width=\linewidth]{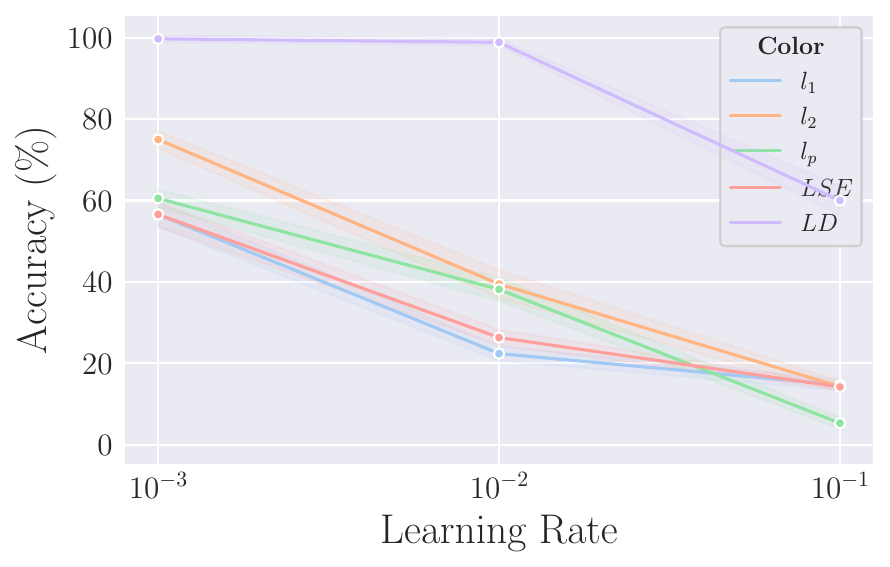}
        \caption{Different learning rate}
        \label{fig:3:lr}
    \end{subfigure}
    \hfill
    \begin{subfigure}[b]{0.49\linewidth}
        \includegraphics[width=\linewidth]{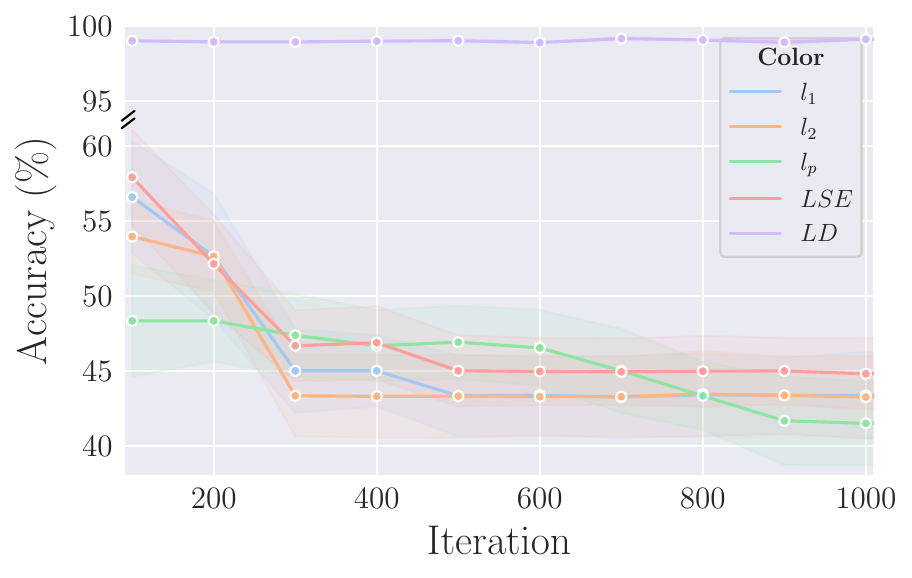}
        \caption{Different iteration}
        \label{fig:3:iteration}
    \end{subfigure}\vspace{-1em}
    \caption{Lineage detection result when finetuning ResNet18 on CIFAR10 with different learning rates and iterations.}
    \label{fig:3}
\end{figure}

\textbf{Learning rate and iteration.} Further analysis was conducted on the effects of different finetuning learning rates and iterations on our methods. This experiment was motivated by the direct impact of learning rate and finetuning iterations on the distance between parent and child models, thus influencing the difficulty of lineage detection. In \cref{fig:3}, we plotted the accuracy of lineage detection for ResNet18 models on CIFAR10 as a function of learning rate and iteration. As the learning rate or iteration increases, lineage detection becomes more challenging, with all methods showing a decreasing tendency in accuracy. However, the learning-based method proves more robust compared to the learning-free ones.

\textbf{Across generation.} Similar to the number of iterations, another factor influencing the difficulty of lineage detection is the number of generations of finetuning. We defined the 7 ResNet18 parent models as the Generation 1 (G1) models and fine-tuned them down 3 generations along  EMNIST-Letters(G2)$\rightarrow$FMNIST(G3)$\rightarrow$EMNIST-Balanced\cite{emnist}(G4). Each generation produced a comparable number of child models to previous experiments. We conduct lineage detection between all descendant-ancestor pairs, with results shown in \cref{tab:mulgen}. It can be observed that the relative advantages of the proposed methods exist under different generation gap sizes, with the learning-based method still performing the best, followed by the learning-free method. However, as the generation gap increases, the general detection performance declines. The red dots in the table indicate the relative accuracy, with darker red representing lower accuracy. The dots in the diagonal blocks have the lightest red, corresponding to the smallest generation gap and highest accuracy, while the darkest dots are in the upper-right block, corresponding to the largest generation gap and lowest accuracy.

\begin{table}[]
\small
\begin{tabular}{@{}l@{$\ \ $}l@{$\ \ $}r@{$\pm$}l@{$\ \ $}r@{$\pm$}l@{$\ \ $}r@{$\pm$}l@{}}
\toprule
                    &                  & \multicolumn{2}{c}{G2} & \multicolumn{2}{c}{G3} & \multicolumn{2}{c}{G4} \\ \midrule
\multirow{5}{*}{G1} & $l_1$ w/o Appx.  & \cb{36} 90.47      & 0.84      & \cb{50} 89.07      & 1.72      & \cb{77} 86.31      & 2.47      \\
                    & CKA w/o Appx.    & \cb{0} 75.59      & 2.19      & \cb{83} 44.44      & 2.89      & \cb{100} 38.09      & 1.15      \\
                    & $l_1$ w Appx.    & \cb{0} 98.92      & 0.59      & \cb{52} 97.02      & 0.51      & \cb{57} 96.83      & 0.82      \\
                    & CKA w Appx.      & \cb{0} 77.38      & 1.93      & \cb{99} 51.59      & 2.98      & \cb{100} 51.19      & 1.27      \\
                    & Lineage Detector & \cb{27} 99.11      & 0.36      & \cb{44} 98.81      & 0.13      & \cb{100} 97.75      & 1.53      \\ \midrule
\multirow{5}{*}{G2} & $l_1$ w/o Appx.  & \multicolumn{2}{c}{-}  & \cb{38} 91.27      & 1.89      & \cb{100} 83.93      & 2.67      \\
                    & CKA w/o Appx.    & \multicolumn{2}{c}{-}  & \cb{36} 65.88      & 2.63      & \cb{21} 67.86      & 2.03      \\
                    & $l_1$ w Appx.    & \multicolumn{2}{c}{-}  & \cb{14} 98.41      & 0.53      & \cb{100} 95.24      & 1.71      \\
                    & CKA w Appx.      & \multicolumn{2}{c}{-}  & \cb{8} 75.41      & 2.81      & \cb{21} 72.03      & 3.17      \\
                    & Lineage Detector & \multicolumn{2}{c}{-}  & \cb{0} 99.61      & 0.83      & \cb{67} 98.38      & 0.02      \\ \midrule
\multirow{5}{*}{G3} & $l_1$ w/o Appx.  & \multicolumn{2}{c}{-}  & \multicolumn{2}{c}{-}  & \cb{0} 94.05      & 1.49      \\
                    & CKA w/o Appx.    & \multicolumn{2}{c}{-}  & \multicolumn{2}{c}{-}  & \cb{24} 66.67      & 1.42      \\
                    & $l_1$ w Appx.    & \multicolumn{2}{c}{-}  & \multicolumn{2}{c}{-}  & \cb{52} 97.02      & 0.75      \\
                    & CKA w Appx.      & \multicolumn{2}{c}{-}  & \multicolumn{2}{c}{-}  & \cb{21} 72.03      & 1.91      \\
                    & Lineage Detector & \multicolumn{2}{c}{-}  & \multicolumn{2}{c}{-}  & \cb{65} 98.41      & 0.98      \\ \bottomrule
\end{tabular}\vspace{-1em}
  \caption{Results of across-generational lineage detection.}
  \label{tab:mulgen}\vspace{-1.5em}
\end{table}

\subsection{Other Tasks and Losses}
\textbf{Dense prediction.} We also investigate the lineage detection performance in object detection and segmentation tasks. For detection, we used 8 models fine-tuned on a subset of the PASCAL~\cite{pascal} dataset as parent models from the officially released DETR~\cite{detr} model, and 56 models further fine-tuned on a non-overlapping subset of PASCAL as child models. For segmentation, we used the officially released DETR panoptic segmentation model as the starting point, similarly fine-tuned on PASCAL across two generations to create parent and child models. \cref{tab:furthersetting} shows proposed methods maintain their advantages.

\textbf{Regularizations.} Finally, we examine the impact of introducing additional regularization losses during finetuning on lineage detection. We considered EWC regularization~\cite{ewc} on parameters and KL-divergence (KLD) regularization~\cite{kd} on outputs, widely used in continual learning and knowledge distillation, respectively. These regularizations represent two settings: EWC directly constrains child model parameters from deviating too far from the parent model parameters, potentially strengthening the lineage relationship and can be considered as a strategy to facilitate lineage detection; KLD requires the fine-tuned model's output to closely resemble another teacher model's output (not the parent), potentially weakening the lineage relationship and can be considered as a strategy to attack lineage detection. The EWC experiment is conducted by adding EWC loss in the original FC+FMNIST setup. The KLD experiment is conducted by introducing an additional teacher model and KL divergence loss on top of the original ResNet18+FMNIST setup. Results in \cref{tab:furthersetting} show that the relative advantages between methods remained largely unchanged, with EWC leading to increased accuracy and KLD to decreased accuracy.

\begin{table}[]
\small
\begin{tabular}{@{}c@{}c@{$\ $}r@{$\pm$}l@{$\ \ $}r@{$\pm$}l@{$\ \ $}r@{$\pm$}l@{$\ \ $}r@{$\pm$}l@{}}
\toprule
                                       &                 & \multicolumn{2}{l}{\scriptsize Detection} & \multicolumn{2}{l}{\scriptsize Segmentation} & \multicolumn{2}{l}{\scriptsize\makecell{ Continual\\Learning}} & \multicolumn{2}{l}{\scriptsize\makecell{Knowledge\\Distillation}} \\ \midrule
\parbox[t]{2mm}{\multirow{5}{*}{\rotatebox[origin=c]{90}{w/o Approx.}}} & $l_1$           &95.86&2.03               &               25.01&3.97                                  & 98.25              & 0.12              & 93.38                & 1.66                \\
                                       & $l_2$           &               91.22&2.26               &                22.91 &2.11                 & 99.74              & 0.38              & 92.92                & 1.59                \\
                                       & $l_{\infty}$    &               78.58&4.01               &                 33.33&4.68                 & 97.81              & 0.54              & 70.65                & 1.21                \\
                                       & $CKA$           &               73.20&4.54               &                 24.98&5.89                 & 39.04              & 1.11              & 44.03                & 3.25                \\
                                       & $DC$            &               71.76&4.81               &                 20.85&4.24                 & 37.84              & 1.02              & 34.43                & 3.01                \\ \midrule
\parbox[t]{2mm}{\multirow{6}{*}{\rotatebox[origin=c]{90}{w Approx.}}}    & $l_1$           &               \underline{96.42}&\underline{1.23}               &                 31.27&3.45                 & 99.32              & 0.12              & 93.57                & 1.75                \\
                                       & $l_2$           &               92.85&1.91               &                 25.01&1.52                 & \textbf{99.83}              & \textbf{0.21}              & \underline{94.03}                & \underline{1.23}                \\
                                       & $l_p$           &               91.07&1.57               &                 \underline{35.41}&\underline{2.29}                 & 99.82              & 0.23              & 82.57                & 2.19                \\
                                       & $LSE$           &               96.23&1.83              &                 22.93&2.31                 & 99.11              & 0.38              & 93.21                & 1.71                \\
                                       & $CKA$           &               78.56&4.43               &                 24.99&5.95                 & 41.23              & 1.11              & 46.79                & 3.38                \\
                                       & $DC$            &               73.21&4.45               &                 27.07&3.84                 & 38.61              & 1.31              & 34.86                & 2.74                \\ \midrule
\multicolumn{2}{l}{\makecell{Lineage\\Detector}}         &               \textbf{99.28}&\textbf{0.72}               &                 \textbf{37.56}&\textbf{4.04}                 & \underline{99.18}              & \underline{0.63}              & \textbf{99.07}                & \textbf{0.07}                \\ \bottomrule
\end{tabular}\vspace{-1em}
  \caption{Lineage detection results for other tasks and losses. The best (the second-best) ones are marked in \textbf{bold} (\underline{underlined}).}
  \label{tab:furthersetting}\vspace{-1.5em}
\end{table}
\vspace{-0.5em}
\section{Conclusion}
In this paper, we propose the task of neural lineage detection, which aims to determine the parent-child relationship between models without relying on external media. To address this task, we first introduce a learning-free, similarity-based detection method that incorporates an approximation of the finetuning process into the model similarity measurement, achieving accuracy beyond directly using similarity measurement. We further propose a transformer-based lineage detector, which shows significant performance gains over the learning-free method.

\section*{Acknowledgment}
This research is supported by the National Research Foundation, Singapore, under its AI Singapore Programme (AISG Award No: AISG2-RP-2021-023).

\clearpage
{
    \small
    \bibliographystyle{ieeenat_fullname}
    \bibliography{main}
}

\addtocontents{toc}{\protect\setcounter{tocdepth}{4}}
\onecolumn
\renewcommand{\thefigure}{S\arabic{figure}}
\renewcommand{\thetable}{S\arabic{table}}
\renewcommand{\thesection}{S\arabic{section}}
\setcounter{figure}{0}
\setcounter{table}{0}
\setcounter{section}{0}
{
    \centering
    \Large
    \textbf{Neural Lineage\\\textit{- Supplementary Material -}}\\
    \vspace{1.0em}
}

\vspace{0.5em}
\allowdisplaybreaks
\begingroup
\hypersetup{linkcolor=black}
\tableofcontents
\endgroup
\clearpage

\section{Further Experiments and Analyses}
\subsection{Learning-Free Methods}
We note two crucial factors controlling the performance of the learning-free methods.
The first is the position of the feature used, and the second is the hyper-parameter $\alpha$. The following two figures present their influences. 

For the feature position, we compare the performance of our method and that of the baseline when using ResNet18 on the DTD, Cifar10, EMNIST-Letters, and FMNIST. The similarity metric is chosen to be the $l_1$-norm, which is generally the best, as indicated by the results in Tab. 1 in the main text. We use the following four features to calculate the similarities, No. 1: the feature after the first convolutional layer, No. 2: the feature after the first ResNet block, No. 3: the feature after the second ResNet block, and No. 4: the feature after the third ResNet block. From No. 1 to No. 4, the feature used to evaluate the similarities gets deeper and deeper. For each feature position, we search for the best $\alpha$. All other configurations are the same as the corresponding original experiments. As shown in \cref{fig:sup1:fp}, although our method consistently outperforms the baseline, the accuracy of both our method and the baseline decreases with increasing feature depth. This trend may be attributed to the fact that as features become deeper, they encapsulate more high-level semantic information. The shallow features of different models might vary significantly, as there could be numerous ways to extract these low-level features while still achieving accurate classification. However, given that the models we selected all have quite high accuracy, their deeper features tend to align more closely with the classification objective. For instance, the mutual information between features and labels increases. Thus, this reduces the differences between the features of different models and leads to a decline in lineage detection accuracy, which results in an accuracy plot that is initially higher but decreases then. 

For $\alpha$, we again compare the performance of our method and that of the baseline when using ResNet18 on the DTD, Cifar10, EMNIST-Letters, and FMNIST. The similarity metric is still $l_1$-norm. We fix the feature position to be the feature after the first ResNet block. All other configurations are the same as the corresponding original experiments. The results are plotted in \cref{fig:sup1:alpha}. The variation in the value of $\alpha$ results in distinct experimental phenomena. With a small $\alpha$, the learning-free method is nothing but the original similarity metric, which leads to no performance gain; with a large $\alpha$, the finetuning approximation is inaccurate, which leads to a significant performance drop; with a proper $\alpha$, the finetuning approximation is valid, our method outperforms the baseline.

\setcounter{figure}{5}
\begin{figure}[h]
    \centering
    \begin{subfigure}[b]{0.3\linewidth}
        \includegraphics[width=\linewidth]{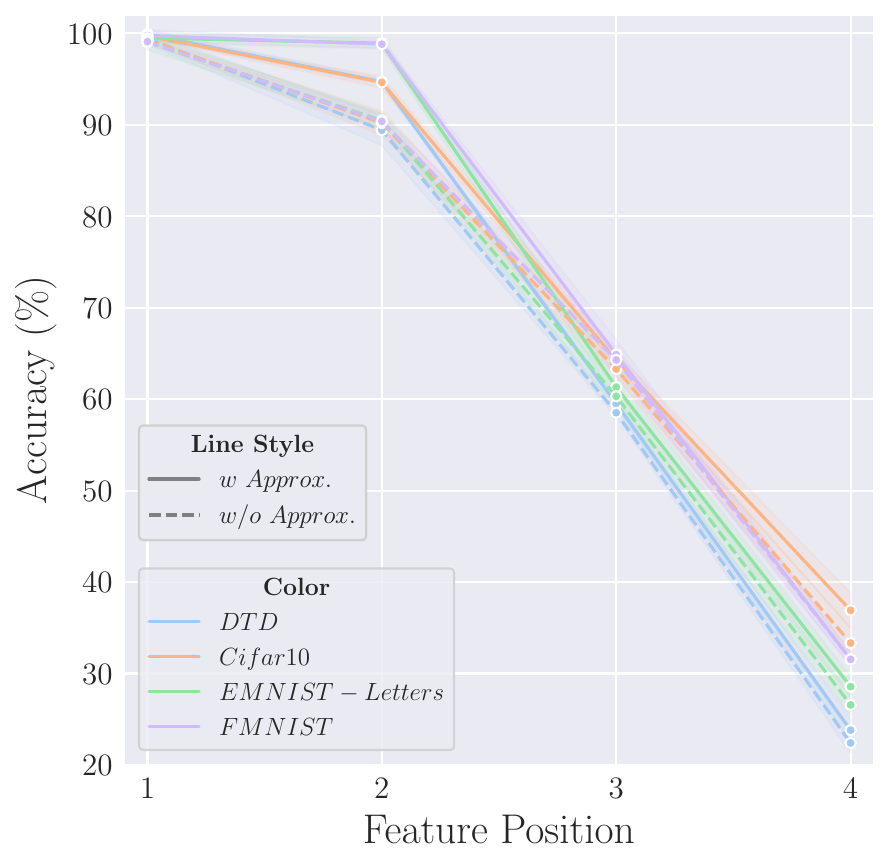}
        \caption{Feature Position}
        \label{fig:sup1:fp}
    \end{subfigure}
    \begin{subfigure}[b]{0.3\linewidth}
        \includegraphics[width=\linewidth]{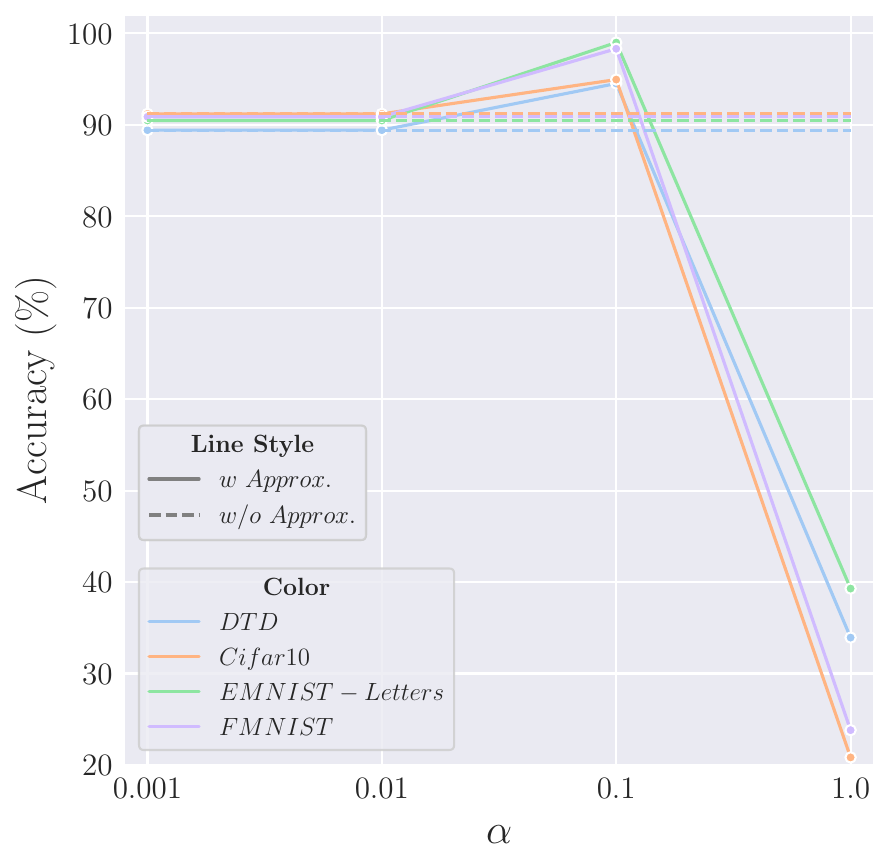}
        \caption{$\alpha$ Value}
        \label{fig:sup1:alpha}
    \end{subfigure}
    \vspace{-1em}
    \caption{Comparison of the performance when feature position and $\alpha$ vary.}
    \label{fig:sup1}
    \vspace{-1em}
\end{figure}

\subsection{Lineage Detector}
\setcounter{table}{3}
In practice, we may encounter scenarios where our parent model set is incomplete, and the child model is not fine-tuned from any model within the parent model set. To investigate this, we design a new experiment based on the ``ResNet18+Cifar10'' and ``ResNet+SVHN'' setups. During dataset construction, we excluded the ``resnet18.tv\_in1k'' model from the parent model set but retained all its fine-tuned child models. This approach resulted in some child models having no known parent model. For child models other than those fine-tuned from ``resnet18.tv\_in1k'', the ground truth labels remain unchanged, pointing to their original parents. However, the child models of ``resnet18.tv\_in1k'' are assigned an additional category to indicate they have ``no parent''.

Consequently, we had six parent models. Using our framework, the lineage detector could output a six-dimensional vector $\bm{s}\in\mathcal{R}^6$ for each child model, representing the similarity scores with all six parent models. To enable the lineage detector to predict the ``no parent'', we concatenate a learnable scalar $s'$ to the end of $\bm{s}$, forming a new vector $[\bm{s}_1;\cdots;\bm{s}_6;s']$, which served as the final logits output of the lineage detector. During parent prediction, if $s'=\max \{\bm{s}_1,\cdots,\bm{s}_6,s'\}$, the model predicts that the child model's parent is not within the existing parent model set.

The rest of the experimental setup remains consistent with the original experiments. The results, presented in \cref{tab:noparent}, show that while no-parent child models can sometimes disrupt lineage detection, such as in the Cifar10 setup, the lineage detector can also maintain a relatively high accuracy under the other setup. Moreover, our model could almost perfectly identify child models whose parent model does not belong to the current parent model set.

\begin{table}[h]
\small
\centering
\begin{tabular}{@{}cr@{$\pm$}lr@{$\pm$}l@{}}
\toprule
                                                & \multicolumn{2}{c}{Cifar10} & \multicolumn{2}{c}{SVHN} \\ \midrule
Overall Accuracy                                & 82.14         & 1.58        & 99.26       & 0.12       \\
Accuracy of the Children Without Real   Parents & 99.65         & 0.37        & 99.92       & 0.21       \\ \bottomrule
\end{tabular}
  \caption{The performance of Lineage Detector when some of the child models do not have real patients.}
  \label{tab:noparent}\vspace{-1.5em}
\end{table}

Benefiting from the structure of the transformer, the lineage detector allows for the use of multiple weights or features as inputs, or it can rely solely on either weights or features. We conducted experiments to analyze the impact of different combinations of weights and features on the performance of the lineage detector. These experiments were carried out under the ``FC+FMNIST'' and ``FC+EMNIST-Letters'' setups, with all other experimental conditions remaining consistent with the original settings, except for the variations in features and weights used. The results are summarized in \cref{tab:mfw}. Comparing the rows of ``Feature after the 1st ReLU Layer'', ``Weight of the 1st Linear Layer'', and the corresponding experiments in the main text Tab. 1, it is evident that the lineage detector, using only features or weights, can surpass the performance of learning-free methods, which use both of them. 
However, comparing the last row with other rows, there is a noticeable performance gap between the lineage detector using either the weight or feature alone and the lineage detector using both of them. We hypothesize that the lineage detector, when provided with both, can learn an approximation of the fine-tuning process similar to Eq. (1) in the main text, thereby enhancing lineage detection effectiveness. Another observation is consistent with the phenomena in \cref{fig:sup1:fp}: deeper features tend to decrease the effectiveness of lineage detection. Integrating features from multiple layers does not mitigate this issue.
\begin{table}[h]
\small
\centering
\begin{tabular}{@{}lr@{$\pm$}lr@{$\pm$}l@{}}
\toprule
                                                                      & \multicolumn{2}{c}{FMNIST} & \multicolumn{2}{c}{EMNIST} \\ \midrule
Feature after the 1st ReLU Layer                                      & 54.58        & 3.31        & 50.01            & 7.37            \\
Feature after the 2nd ReLU Layer                                      & 9.02         & 2.86        & 14.86            & 3.93            \\
Feature after the 3rd ReLU Layer                                      & 9.08         & 3.08        & 12.75            & 2.06            \\
Feature after the 1st ReLU Layer +   Feature after the 2nd ReLU Layer & 40.92        & 2.28        & 45.56            & 5.09            \\ \midrule
Weight of the 1st Linear Layer                                        & 57.34        & 4.97        & 62.94            & 5.81            \\
Weight of the 2nd Linear Layer                                        & 76.98        & 4.09        & 91.24            & 1.99            \\
Weight of the 3rd Linear Layer                                        & 89.11        & 2.51        & 89.23            & 2.02            \\
Weight of the 1st Linear Layer + Weight   of the 2nd Linear Layer     & 66.32        & 6.01        & 87.91            & 4.03            \\ \midrule
Feature after the 1st ReLU Layer + Weight   of the 1st Linear Layer   & \textbf{97.86}        & \textbf{1.19}        & \textbf{93.61}            & \textbf{2.37}            \\ \bottomrule
\end{tabular}
  \caption{The performance of Lineage Detector when different weight and feature combination is used.}
  \label{tab:mfw}\vspace{-1.5em}
\end{table}

\subsection{Average Performance Across Datasets}
\begin{table*}[]
\small
\begin{tabular}{@{}l@{$\quad$}r@{$\pm$}l@{$\quad$}r@{$\pm$}l@{$\quad$}r@{$\pm$}l@{$\quad$}r@{$\pm$}l@{$\ |\ $}r@{$\pm$}l@{$\quad$}r@{$\pm$}l@{$\quad$}r@{$\pm$}l@{$\quad$}r@{$\pm$}l@{$\quad$}r@{$\pm$}l@{}}
\toprule
                        & \multicolumn{8}{c}{Without Approximation}                                                                    & \multicolumn{8}{c}{With Approximation}                                                                       & \multicolumn{2}{c}{\multirow{2}{*}{\makecell{Lineage\\ Detector}}} \\ \cmidrule(lr){2-17}
                        & \multicolumn{2}{c}{$l_1$} & \multicolumn{2}{c}{$l_2$} & \multicolumn{2}{c}{$CKA$} & \multicolumn{2}{c}{$DC$} & \multicolumn{2}{c}{$l_1$} & \multicolumn{2}{c}{$l_2$} & \multicolumn{2}{c}{$CKA$} & \multicolumn{2}{c}{$DC$} & \multicolumn{2}{c}{}                                    \\ \midrule
FCNet & 63.41        & 1.27       & 65.03        & 1.41       & 26.22        & 1.18       & 26.65       & 1.19       & 64.95        & 1.49       & \underline{66.66}        & \underline{1.44}       & 32.27        & 1.25       & 32.17       & 1.41       & \textbf{98.29}                       & \textbf{0.90}                      \\
ResNet18                & 92.72        & 0.91       & 91.67        & 0.83       & 64.54        & 1.82       & 56.23       & 1.95       & \underline{97.31}        & \underline{0.60}       & 92.52        & 0.92       & 66.86        & 2.12       & 56.99       & 1.94       & \textbf{99.30}                       & \textbf{0.23 }                     \\ \bottomrule
\end{tabular}
\caption{The average performance of various methods across different datasets in Tab. 1 of the main text. The best (the second-best) ones are marked in \textbf{bold} (\underline{underlined}).}
  \label{tab:avaerage_dataset}
\end{table*}
To facilitate the comparison of the performance of various methods, we calculate the average performance of each method across classification datasets. The results are shown in \cref{tab:avaerage_dataset}.
The learning-based method maintaining the highest accuracy. Among the learning-free methods, $l_1$ shows superiority for ResNet, while $l_2$ excels for Fully Connected Network. Norm distance consistently outperforms advanced representation similarity. 

\section{Implementation Details}
Experiments are conducted using Nvidia RTX 3090 if no special specification.
\subsection{Classification Model Experiments in Table 1}
\subsubsection{Parent and Child Models}
We employ two families of classification models: one comprising three-layer fully connected networks only with ReLU as the nonlinear activation, and the other is the ResNet18. 

For experiments using a fully connected network on FMNIST and EMNIST-Letters, the input and output dimensions of the three-layer fully connected network models are adjusted according to the dataset, with hidden feature sizes being fixed to 1024, 256, and 128. We initially train 20 parent models on MNIST, each with different training configurations. Their batch sizes are one of \{64, 256, 1024\}, and their learning rates are one of \{0.1, 0.01, 0.001\}. We use four different random seeds. The training epochs are set to 50. We use the Adam optimizer to train. From the 36 trained models, we select the top 20 with the highest test accuracy as our parent models for experiments. Starting from these 20 parent models, we fine-tune them on FMNIST and EMNIST-Letters datasets. The fine-tuning learning rates are one of \{0.01, 0.001, 0.0001\}, and batch sizes are one of \{1024, 256\}. We again use four different random seeds. The tuning epochs are set to 36. We use the Adam optimizer. Eventually, on each dataset, we obtain 480 models and select those with test accuracy above 80\% as our child models for subsequent experiments. This results in 228 child models each for FMNIST and EMNIST-Letters.

For experiments using a fully connected network on Cifar10, SVHN, and variants of Cifar10, the input and output dimensions of the three-layer fully connected network models are adjusted according to the dataset, with hidden feature sizes being fixed to 1024, 256, and 128. We initially train 12 parent models on Cifar100, each with different training configurations. Their batch sizes are either 256 or 1024, and their learning rates are either 0.01 or 0.001. We use three different random seeds. The training epochs are set to 50. We use the Adam optimizer to train. We select all of them as the parent models, as their testing accuracy are all above 80\%. Starting from these 12 parent models, we fine-tune on Cifar10, SVHN, and Cifar10 variants. The fine-tuning hyperparameters are similar to the previous setup, with learning rates of 0.01, 0.001, or 0.0001, and batch sizes of 1024 or 256, using four different random seeds. The training epochs are set to 20, using the Adam optimizer. Eventually, on each dataset, we obtain 288 models, selecting those with test accuracy above 80\% as our child models for subsequent experiments. This resulted in 254, 264, 132, 264, and 260 child models for Cifar10, SVHN, Cifar10-5shot, Cifar10-50shot, and Cifar10-imbalanced datasets, respectively.

For the experiments using a ResNet18 network, we use 7 ResNet18 networks in the timm~\cite{timm} package as the parent models. All of the networks are trained on the ImageNet dataset. The networks we used are ``resnet18.a1\_in1k'', ``resnet18.a2\_in1k'', ``resnet18.a3\_in1k'', ``resnet18.gluon\_in1k'', ``resnet18.fb\_ssl\_yfcc100m\_ft\_in1k'', ``resnet18.fb\_swsl\_ig1b\_ft\_in1k'' and ``resnet18.tv\_in1k''. If the finetuning dataset is FMNIST or EMNIST-Letters, the kernel of the first convolutional layer is replaced by its average along the channel dimension to support grayscale input. We fine-tune these 7 models on the datasets listed in Tab. 1. The fine-tuning hyperparameters are similar to the previous setups, with learning rates of 0.01, 0.001, or 0.0001, and batch sizes of 1024 or 256, using four different random seeds. The training epochs are set to 20. The optimizer is Adam. The head of the network is modified according to the number of categories of the target dataset and reinitialized. Eventually, on each dataset, we obtain 168 models and select those with test accuracy above 80\% as our child models for subsequent experiments. For each of FMNIST, EMNIST-Letters, Cifar10, SVHN, DTD, Pets, and Cifar10-imbalanced datasets, we obtain 168 child models. For Cifar10-5shot and Cifar10-50shot, we obtain 90 and 94 child models respectively.

\subsubsection{Our Methods and Baselines}
When the fully connected network is used, both our learning-free methods and the baselines use the output feature of the first ReLU layer to calculate the similarity.  When ResNet18 is used, both our learning-free methods and the baselines use the output feature of the first ResNet block to calculate the similarity. Though our learning-based lineage detector is able to use only the feature as input, for a fair comparison, we use both the weight and the feature as input in this set of experiments.  When the fully connected network is used, the weight of the first linear layer and the output feature of the first ReLU layer are taken as the input of the lineage detector; when ResNet18 is used, the weight of the last linear layer in the first ResNet block and the output feature of the first ResNet block are taken as the input of the lineage detector. For the learning-free method, the hyper-parameter $\alpha$ is searched within \{0.001, 0.01, 0.1\}. For all the methods, the number of samples to evaluate the similarity is set to 512.
The parameter $p$ in $p$-norm is always set to be 4. The parameter $t$ in log-sum-exp is always set to 0.01.

\subsection{Classification Model Experiments in Table 2}
In this experiment, we evaluated our method's capability for cross-generational detection across four generations of models. For the first vs. second-generation experiment, the parent and child models are those used in the ResNet+EMNIST-Letters experiment in Tab. 1. We refer to the first-generation models as root models, with all subsequent models being their descendants. For each root model, we select the second-generation child model with the highest test accuracy as the parent model for the second vs. third-generation experiment. These models are fine-tuned on the FMNIST dataset to obtain the third-generation models, which served as child models in the second vs. third-generation experiment. Similarly, for each root model, we select the third-generation descendant with the highest test accuracy as the parent model for the third vs. fourth-generation experiment. These models are fine-tuned on the EMNIST-Balanced dataset to obtain the fourth-generation models, which are the child models in the third vs. fourth-generation experiment. The fine-tuning hyperparameters are consistent with previous setups, including learning rates of 0.01, 0.001, or 0.0001, batch sizes of 1024 or 256, and four different random seeds. The training epochs are set to 20, using the Adam optimizer, and the network head is modified and reinitialized according to the number of categories in the target dataset. Ultimately, we obtained 168 models for each dataset, selecting only those with test accuracy above 80\% for further experiments.

Through this approach, we constructed a model family comprising 7 first-generation models; 168 second-generation models, all of which are child models in the first vs. second-generation experiment, with 7 serving as parent models in the second vs. third-generation experiment; 127 third-generation models, all child models in the second vs. third-generation experiment, with 7 serving as parent models in the third vs. fourth generation experiment; and 168 fourth-generation models, all child models in the third vs. fourth generation experiment. For the first vs. third and first vs. fourth-generation experiments, we used the 7 root models as parent models and the third and fourth-generation models as child models. For the second vs. fourth-generation experiment, the 7 parent models from the second vs. third-generation experiment were used as parent models, with the fourth-generation models as child models.

The implementation of our methods and the baselines is consistent with the settings used in the ResNet+EMNIST-Letters experiment in Tab. 1.

\subsection{Detection and Segmentation Model Experiments}
\subsubsection{Parent and Child Models}
For the detection and segmentation experiments, we utilized the official DETR+ResNet50 model released for detection and segmentation, respectively. We first created two subsets of the PASCAL dataset: SubSet 1, containing all categories except animals, and SubSet 2, comprising all animal categories. For detection, we fine-tune the official ``DETR\_R50\_detection'' model on SubSet 1. The fine-tuning learning rate is either 0.005, 0.001, 0.0005, or 0.0001, the batch size is 16, and the epoch is 25. We use two different random seeds. All other experimental settings follow the default training configurations of DETR. This process yielded 8 models. We use all of them as parent models. These parent models are further fine-tuned on SubSet 2. The fine-tuning learning rate is either 0.005, 0.001, 0.0005, or 0.0001, the batch size is 16, and the epoch is 25. We use four different random seeds. This resulted in 128 models, from which we select those with training accuracy above 80\%, resulting in 56 child models.

For the segmentation experiment, we fine-tune the official ``DETR\_R50\_segmentation'' model on SubSet 1. The fine-tuning learning rate is either 0.005, 0.001, 0.0005, or 0.0001, the batch size is 16, and the epoch is 25. We use two different random seeds. All other experimental settings follow the default training configurations of DETR. This process yielded 8 models. We use all of them as parent models. These parent models are further fine-tuned on SubSet 2. The fine-tuning learning rate is either 0.005, 0.001, 0.0005, or 0.0001, the batch size is 16, and the epoch is 25. We use four different random seeds. This resulted in 128 models, from which we select those with training accuracy above 80\%, resulting in 49 child models.

\subsubsection{Our Methods and Baselines}
For both the detection and segmentation, both our learning-free methods and the baselines use the output feature of the first transformer block in the encoder of DETR.
For both the detection and segmentation, the weight of the last linear layer in the first transformer block in the encoder of DETR and the output feature of the first transformer block in the encoder of DETR are taken as the input of the lineage detector. 
For the learning-free method, the hyper-parameter $\alpha$ is searched within \{0.001, 0.01, 0.1\}. For all the methods, the number of samples to evaluate the similarity is set to 16.
The parameter $p$ in $p$-norm is always set to be 4. The parameter $t$ in log-sum-exp is always set to 0.01.

\subsection{Continual Learning and Knowledge Distillation Experiments}
We investigate the impact of two types of loss functions with additional regularization terms on lineage detection: Elastic Weight Consolidation (EWC) from Continual Learning and Kullback-Leibler Divergence (KLD) from Knowledge Distillation. 

The EWC experiment involves incorporating EWC loss into the original FC+FMNIST setup. The implementation of EWC loss adheres to the methodology outlined in the original paper.~\cite{ewc} The weight of the EWC loss is set at 10, 20, or 100. The number of random seeds is reduced to two. The selection of parent models, other settings for tuning child models, and the method for filtering child models remain consistent with the FC+FMNIST setup. Ultimately we obtain 227 child models.

In the KLD experiment, an additional teacher model and KL divergence loss are introduced alongside the original ResNet18+FMNIST setup. This experiment utilizes ``resnet18.a1\_in1k'', ``resnet18.a2\_in1k'', ``resnet18.a3\_in1k'', ``resnet18.gluon\_in1k'',  ``resnet18.fb\_swsl\_ig1b\_ft\_in1k'' and ``resnet18.tv\_in1k'' as parent models. The child model with the highest test accuracy fine-tuned from the ``resnet18.fb\_ssl\_yfcc100m\_ft\_in1k'' model on FMNIST is used as the teacher model. During the tuning of child models from parent models, a KLD term is added, requiring the final output of the child models to be similar in KLD to the final output of the teacher model. This approach is in line with standard response-based knowledge distillation practices.~\cite{kd_survey} The weight of the KLD loss is set to 1, with the temperature parameter in the KLD loss adjusted to 1.0, 2.0, or 5.0. The number of random seeds is again reduced to two. Other settings for training child models and the method for filtering child models are consistent with the ResNet18+FMNIST setup, resulting in 168 child models.

In the continual learning experiment, the implementation of our methods and the baselines is consistent with the settings used in the FC+FMNIST experiment in Tab. 1. In the knowledge distillation experiment, the implementation of our methods and the baselines is consistent with the settings used in the  ResNet18+FMNIST experiment in Tab. 1.

\subsection{Lineage Detector}
\subsubsection{Dataset for Lineage Detector}
The prerequisites of training and utilizing the lineage detector involve constructing an appropriate dataset. Assuming no constraints on sample size, a dataset suitable for training and testing the lineage detector can be constructed from each of our "neural network architecture + finetuning dataset" settings. Taking the "FC+FMNIST" setup as an example, we illustrate how to construct a dataset for the lineage detector, applicable similarly to other settings. We assume that the input weights for the lineage detector are all parameters of the neural network except for the head, and the input feature is the backbone output, \emph{i.e.}, the input feature of the neural network head. If one opts not to use weights or features as inputs, they can be set as zero matrices, and the corresponding encoder in the lineage detector can be omitted.

Let $f_p:\mathcal{X}\rightarrow\mathcal{Y}$ and $f_c:\mathcal{X}\rightarrow\mathcal{Y}$ denote the parent neural network and the child neural network parameterized by $\bm{\theta}_p$ and $\bm{\theta}_c$, respectively. Suppose there are $M$ parent models $\{f_p^{(m)}\}_{m=1}^M$ and $M_c$ child models $\{f_c^{(m)}\}_{m=1}^{M_c}$. Suppose there are $N$ FMNIST samples $\{(\bm{x}^{(i)},\bm{y}^{(i)})\}_{i=1}^N$. Let $\bm{X}$ denote the images in batch format. Let $\tilde{\bm{\theta}}$ denote the parameters except the parameters in the head. Let $\tilde{f}$ denote the neural network without the head.

First, we collect the parameters $\tilde{\bm{\theta}}$ of all parent and child models, 
$\{\tilde{\bm{\theta}}_p^{(m)}\}_{m=1}^M$ and $\{\tilde{\bm{\theta}}_c^{(m)}\}_{m=1}^{M_c}$. Second, we evaluate the backbone output of parent and child models, $\{\Tilde{f}_p^{(m)}(\bm{X})\}_{m=1}^M$ and $\{\Tilde{f}_c^{(m)}(\bm{X})\}_{m=1}^{M_c}$. For each of the $\tilde{\bm{\theta}}_p^{(m)}$, $\tilde{\bm{\theta}}_c^{(m)}$, $\Tilde{f}_p^{(m)}(\bm{X})$, and $\Tilde{f}_c^{(m)}(\bm{X})$ obtained, we vectorize it and reshape them into a matrix, such that, $\tilde{\bm{\theta}}_p^{(m)}\in\mathcal{R}^{H_{\theta}\times W_{\theta}}$, $\tilde{\bm{\theta}}_c^{(m)}\in\mathcal{R}^{H_{\theta}\times W_{\theta}}$, $\Tilde{f}_p^{(m)}(\bm{X})\in\mathcal{R}^{H_{F}\times W_{F}}$, and $\Tilde{f}_c^{(m)}(\bm{X})\in\mathcal{R}^{H_{F}\times W_{F}}$, where $W_{\theta}$, $H_{\theta}$, $W_{F}$, and $H_{F}$ indicate the shape of the matrix.

With $\{\tilde{\bm{\theta}}_p^{(m)}\}_{m=1}^M$, $\{\Tilde{f}_p^{(m)}(\bm{X})\}_{m=1}^M$, $\tilde{\bm{\theta}}_c^{(n)}$ and $\Tilde{f}_c^{(n)}(\bm{X})$ we can generate the probabilities of the parent models for the child model $n$. 
Thus, in the dataset for lineage detector, each sample contains five parts, the parent weights $\{\tilde{\bm{\theta}}_p^{(m)}\}_{m=1}^M$, the parent features $\{\Tilde{f}_p^{(m)}(\bm{X})\}_{m=1}^M$, the weight of a specific child model $\tilde{\bm{\theta}}_c$, the feature of a specific child model  $\Tilde{f}_c(\bm{X})$ and the index of the real parent of this child model as the ground truth label. The number of samples in a dataset for lineage detector is the number of child models.

Based on this dataset, once the partition of training and test sets is completed, training and testing can proceed. 

\subsubsection{Model and Training Configuration}
Both the weight encoder and the feature encoder are two-layer convolutional neural networks. The number of the input channels of the first layer is 2. The number of the output channels of the first layer is set to 3. The number of the output channels of the second layer is set to 32. The kernel size of the first layer is 1. The kernel size of the second layer is 3. The transformer detector encompasses one transformer layer with all default settings in PyTorch.

The training learning rate is 0.01, the batch size is 1, and the epoch is 100 for all experiments. The Adam optimizer is used.

\subsection{Sub-Figure 1b}
This visualization represents the experimental results of the classification models discussed in the main text. The figure encompasses four generations of models. We select ``resnet18.a1\_in1k'', ``resnet18.a2\_in1k'', ``resnet18.a3\_in1k'', ``resnet18.gluon\_in1k'' and ``resnet18.tv\_in1k'' as the root models. These five models are all used in our experiments of the ResNet in the main text. We mark these five models as the models of the first generation. Models fine-tuned on EMNIST-Letters from these root models constitute the second generation. The third generation comprises models fine-tuned on FMINST from the second generation, and the fourth generation includes models fine-tuned on EMNIST-Balanced from the third generation.  The total number of models is 226. Due to space constraints, not all models used in our experiments are depicted; only the five root models and their descendants are included in this visualization.

The diagram features three concentric rings. The innermost ring consists of multiple small arches, each representing a neural network. The arches vary in width, with the width being proportional to the number of descendants of that model. The middle ring has 20 arches, divided into five color groups, each color representing a different root model. Models covered by the arches with the same color share the same root model, indicating a common ancestor. The outer ring is segmented into four colors, corresponding to the generation of the models; models covered by the arches with the same color belong to the same generation.

At the center of the diagram, arcs with three different colors represent our predictions of the model lineage, with each color indicating a different type of ancestor, such as parent, grandparent, or great-grandparent. Our lineage detector accurately predicts the lineage relationships of all model pairs. Some arcs in the diagram have increased transparency solely for visual clarity and do not carry additional meaning.
\subsection{Sub-Figure 1c}
We utilized nine ViT-Base models from the timm library to construct our hybrid model. The name of these nine models in the timm are: ``vit\_base\_patch16\_224.augreg\_in21k\_ft\_in1k'', ``vit\_base\_patch32\_224.sam\_in1k'', ``vit\_base\_r50\_s16\_224.orig\_in21k'', ``vit\_base\_patch8\_224.augreg2\_in21k\_ft\_in1k'', ``vit\_base\_patch16\_224.mae'', ``vit\_base\_patch32\_clip\_224.openai\_ft\_in1k'', and ``vit\_base\_patch32\_clip\_224.laion2b\_ft\_in12k\_in1k''. The first 6 models are released by Google, the 7th model is released by Meta, and the last 2 models are released by OpenAI. We construct the hybrid model by the following steps. First, starting with the model ``vit\_base\_patch16\_224.augreg\_in21k\_ft\_in1k'', we replace its third transformer layer with the third transformer layer from the model ``vit\_base\_patch32\_224.sam\_in1k''. Now we have a hybrid model. Continuously, we replace the fourth transformer layer of the hybrid model with the fourth transformer layer from the model ``vit\_base\_r50\_s16\_224.orig\_in21k''. We keep doing this until the third transformer layer to the tenth transformer layer are all replaced by the corresponding layers in another model. The final hybrid model is shown in the SubFig. 1c.  We then reinitialized the linear head of this hybrid model and fine-tuned it on the CIFAR-10 dataset. The fine-tuning process used a learning rate of 1e-4, a batch size of 128, and epochs 30, with the Adam optimizer. Prior to fine-tuning, the hybrid model had an accuracy of 0\%, which increased to 100\% on the training set and 45\% on the test set after fine-tuning.

Next, we tend to predict the origin of the transformer layers in the hybrid model. For the lineage detection of the i-th transformer layer, the sub-network of the hybrid model up to the i-th transformer layer is regarded as the child model. The candidate parent models are constructed as follows. We concatenate the sub-network of the hybrid model before the i-th transformer layer with the i-th transformer layer of each of the nine parent models, resulting in nine new networks. These networks had the same input as the original ViT-Base model and their outputs have the sample shape as output of  the i-th transformer layer of the original ViT model. We fed the same data into these nine networks to obtain their outputs, which served as the parent feature inputs for our lineage detector. The features of the i-th transformer layer of the hybrid model, under the same input, are used as the child feature inputs for the lineage detector. In this task, we only use features without any network weights as the input of the lineage detector. For efficiency, we used a sample size of 128 and only the first 64 tokens of the feature. Because there is not enough data in this ViT setting to train the lineage detector. We turn to another setting we used. We use the parent and child models in the ``FC + FMNIST'' setting.
We use the output feature of the first ReLU of the fully connected network as the training data to train the lineage detector.

There is another detail worth discussing. Because the probability distribution of the softmax function can be modified by scaling the input of the softmax function. With a large enough scaling parameter, the probability distribution can be made close to a one-hot vector, which might appear to offer high certainty in predictions and indicate the superior performance of the model. However, we believe this to be misleading and not truly reflective of the model's predictive output. Therefore, before inputting into the softmax, we normalized all values by correcting their mean to be zero and dividing them by the number of parent models 9. Such adjustment does not change the prediction and the sorting results of the probability output of the softmax function, only influences the value of the probability.
\subsection{Figure 2}
The experiment is conducted on the Cifar10 dataset. The number of samples used to calculate the similarity is 10. To show the influence of the number of parameters, we fix the output feature size to be 5 and compare ResNet18, ResNet34, and ResNet50, whose number of parameters are roughly 12M, 22M, and 25M respectively. To show the influence of output feature size, we fix the network to be ResNet18 and change its output dimension to 5, 10 or 15 by modifying its final fc layer. To avoid mutual interference between experiments, the experiments are conducted one by one. The server used to conduct the experiments has one RTX 2080 Ti and 16-core Intel(R) Xeon(R) Platinum 8255C CPU.
\subsection{Sub-Figure 3a}
In this experiment, we calculate the similarity between one parent model and its 20 child models using the step-by-step method and the proposed approximation. The network architecture is a three-layer fully connected network only with ReLU as the non-linear activation. The parent model is trained on MNIST dataset with a learning rate of 0.01 and a batch size of 64. The child models are fine-tuned on the FMNIST dataset with a learning rate of 0.001. The finetuning batch size is one in \{1024, 256\}. The finetuning iteration is one in \{100, $\cdots$, 1000\}. The number of samples used to calculate the similarity is 10. The number of output features is set to 10.
\subsection{Sub-Figure 3b}
In this experiment, we compare the distribution of the change of the similarity after the finetuning process is taken into consideration. We use a three-layer fully connected network only with ReLU as the non-linear activation as the real parent model. This model is trained on the MNIST dataset with a learning rate of 0.01 and a batch size of 64. All the child models are fine-tuned from this model. In total, we use 24 child models. They are fine-tuned on the FMNIST dataset. The finetuning batch size is one in  \{1024, 256\}. The finetuning learning rate is one in  \{0.1,0.01,0.001\}. The finetuning iteration in one in \{100, 200\}. We also use two different random seeds. We train another model on MNIST dataset with a learning rate of 0.01 and a batch size of 1024 as the fake parent model. The number of samples used to calculate the similarity is 5000. The number of output features is set to 10. For each parent-child model pair, no matter whether the parent is fake or real, we first calculate the similarity without the approximation of the finetuning process, then calculate the similarity with the approximation of the finetuning process using our proposed method, and finally, we evaluate the change in similarity by subtracting the former value form the latter one. The extreme values are omitted when plotting the figure.
\section{Derivations}
\label{sec:derivations}
This section includes the derivations for the Proposition 1 in the main text. Before the derivations, we introduce some notations to simplify the formulation and ease the reading. Let $f(\mathcal{X})\in\mathbb{R}^{N\times K}$ denote the output of the neural network in the matrix form taking the entire dataset as the input with each row corresponding to output for one data sample. Specifically, 
\begin{align*}
    \bm{X}&\triangleq f_p(\mathcal{X}) \triangleq [f_p(\bm{x}^{(1)}),\cdots,f_p(\bm{x}^{(N)})]^T,\\
    \bar{\bm{X}}&\triangleq \bar{f}_p(\mathcal{X}) \triangleq [\bar{f}_p(\bm{x}^{(1)}),\cdots,\bar{f}_p(\bm{x}^{(N)})]^T,\\
    \bm{Y}&\triangleq f_c(\mathcal{X}) \triangleq [f_c(\bm{x}^{(1)}),\cdots,f_c(\bm{x}^{(N)})]^T,\\
    \bm{G}&\triangleq\nabla_{\bm{\theta}_p}f_p(\mathcal{X})(\bm{\theta}_c-\bm{\theta}_p)\triangleq[\nabla_{\bm{\theta}_p}f_p(\bm{x}^{(1)})(\bm{\theta}_c-\bm{\theta}_p),\cdots,\nabla_{\bm{\theta}_p}f_p(\bm{x}^{(N)})(\bm{\theta}_c-\bm{\theta}_p)]^T
\end{align*}
where the single letter $\bm{X}$, $\bm{\bar{X}}$, $\bm{Y}$, and $\bm{G}$ is to further simplify the notation. For a matrix $\bm{A}\in\mathbb{R}^{a\times b}$, let $\bm{A}_{:k}\in\mathbb{R}^{a\times 1}$, $\bm{A}_{i:}\in\mathbb{R}^{1\times b}$, and $\bm{A}_{ik}\in\mathbb{R}$ denote the $k$-th column, the $i$-th row, and the element at $i$-th row and $k$-th column. Let $\delta$ denote the high-order terms, which are omitted. Let $\bm{1}$ denote the all-one column vector with proper dimension. Let $\bm{I}$ denote the identity matrix with proper dimension. Let $\bm{H}\triangleq \bm{I} - \frac{1}{N}\bm{1}\bm{1}^T$ denote the centering matrix,

\subsection{Linear Approximation of $L_1$ Norm}
\begin{align}
    s(\Bar{f}_p, f_c)&=-\frac{1}{NK}\sum_{i=1}^N||\bm{\Bar{X}}_{i:}- \bm{Y}_{i:}||_1 \\
    &=-\frac{1}{NK}\sum_{i=1}^N||\bm{X}_{i:}+ \bm{G}_{i:}- \bm{Y}_{i:}||_1 \\
    &=-\frac{1}{NK}\sum_{i=1}^N\sum_{k=1}^K|\bm{X}_{ik}+ \bm{G}_{ik}- \bm{Y}_{ik}|\\
    &\approx-\frac{1}{NK}\sum_{i=1}^N\sum_{k=1}^K[|\bm{X}_{ik}-\bm{Y}_{ik}| + \sign(\bm{X}_{ik}-\bm{Y}_{ik})\bm{G}_{ik}] + \delta \\
    &=-\frac{1}{NK}\sum_{i=1}^N\sum_{k=1}^K|\bm{X}_{ik}-\bm{Y}_{ik}| -\frac{1}{NK}\sum_{i=1}^N\sum_{k=1}^K \sign(\bm{X}_{ik}-\bm{Y}_{ik})\bm{G}_{ik} + \delta \\
    &=s(f_p, f_c) -\frac{1}{NK}\sum_{i=1}^N\sum_{k=1}^K \sign(\bm{X}_{ik}-\bm{Y}_{ik})\bm{G}_{ik} + \delta \\
    &=s(f_p, f_c) -\frac{1}{NK}\sum_{i=1}^N\sum_{k=1}^K \sign(\bm{X}_{ik}-\bm{Y}_{ik})\nabla_{\bm{\theta}_p}\bm{X}_{ik}(\bm{\theta}_c-\bm{\theta}_p) + \delta \\
    &=s(f_p, f_c) -\frac{1}{NK}\nabla_{\bm{\theta}_p}\big[\sum_{i=1}^N\sum_{k=1}^K \sg[\sign(\bm{X}_{ik}-\bm{Y}_{ik})]\bm{X}_{ik}\big](\bm{\theta}_c-\bm{\theta}_p) + \delta \\
    &=s(f_p, f_c) -\frac{1}{NK}\nabla_{\bm{\theta}_p}\langle\sg[\sign(\bm{X}-\bm{Y})],\bm{X}\rangle(\bm{\theta}_c-\bm{\theta}_p) + \delta
\end{align}
\subsection{Linear Approximation of $L_2$ Norm}
\begin{align}
    s(\Bar{f}_p, f_c)&=-\frac{1}{NK}\sum_{i=1}^N||\bm{\Bar{X}}_{i:}- \bm{Y}_{i:}||_2^2 \\
    &=-\frac{1}{NK}\sum_{i=1}^N||\bm{X}_{i:}+ \bm{G}_{i:}- \bm{Y}_{i:}||_2^2 \\
    &=-\frac{1}{NK}\sum_{i=1}^N\sum_{k=1}^K(\bm{X}_{ik}+ \bm{G}_{ik}- \bm{Y}_{ik})^2\\
    &\approx-\frac{1}{NK}\sum_{i=1}^N\sum_{k=1}^K[(\bm{X}_{ik}-\bm{Y}_{ik})^2 + 2(\bm{X}_{ik}-\bm{Y}_{ik})\bm{G}_{ik}] + \delta \\
    &=-\frac{1}{NK}\sum_{i=1}^N\sum_{k=1}^K(\bm{X}_{ik}-\bm{Y}_{ik})^2 -\frac{1}{NK}\sum_{i=1}^N\sum_{k=1}^K 2(\bm{X}_{ik}-\bm{Y}_{ik})\bm{G}_{ik} + \delta \\
    &=s(f_p, f_c) -\frac{1}{NK}\sum_{i=1}^N\sum_{k=1}^K 2(\bm{X}_{ik}-\bm{Y}_{ik})\bm{G}_{ik} + \delta \\
    &=s(f_p, f_c) -\frac{1}{NK}\sum_{i=1}^N\sum_{k=1}^K 2(\bm{X}_{ik}-\bm{Y}_{ik})\nabla_{\bm{\theta}_p}\bm{X}_{ik}(\bm{\theta}_c-\bm{\theta}_p) + \delta \\
    &=s(f_p, f_c) -\frac{2}{NK}\nabla_{\bm{\theta}_p}[\sum_{i=1}^N\sum_{k=1}^K \sg(\bm{X}_{ik}-\bm{Y}_{ik})\bm{X}_{ik}](\bm{\theta}_c-\bm{\theta}_p) + \delta \\
    &=s(f_p, f_c) -\frac{2}{NK}\nabla_{\bm{\theta}_p}\langle\sg(\bm{X}-\bm{Y}),\bm{X}\rangle(\bm{\theta}_c-\bm{\theta}_p) + \delta
\end{align}
\subsection{Linear Approximation of $L_p$ Norm}
\begin{align}
    s(\Bar{f}_p, f_c)&=-\frac{1}{NK}\sum_{i=1}^N||\bm{\Bar{X}}_{i:}- \bm{Y}_{i:}||_p^p \\
    &=-\frac{1}{NK}\sum_{i=1}^N||\bm{X}_{i:}+ \bm{G}_{i:}- \bm{Y}_{i:}||_p^p \\
    &=-\frac{1}{NK}\sum_{i=1}^N\sum_{k=1}^K|\bm{X}_{ik}+ \bm{G}_{ik}- \bm{Y}_{ik}|^p\\
    &\approx-\frac{1}{NK}\sum_{i=1}^N\sum_{k=1}^K[|\bm{X}_{ik}-\bm{Y}_{ik}|^p + p|\bm{X}_{ik}-\bm{Y}_{ik}|^{p-1}\sign(\bm{X}_{ik}-\bm{Y}_{ik})\bm{G}_{ik}] + \delta \\
    &=-\frac{1}{NK}\sum_{i=1}^N\sum_{k=1}^K|\bm{X}_{ik}-\bm{Y}_{ik}|^p -\frac{1}{NK}\sum_{i=1}^N\sum_{k=1}^K p|\bm{X}_{ik}-\bm{Y}_{ik}|^{p-1}\sign(\bm{X}_{ik}-\bm{Y}_{ik})\bm{G}_{ik} + \delta \\
    &=s(f_p, f_c) -\frac{1}{NK}\sum_{i=1}^N\sum_{k=1}^K p|\bm{X}_{ik}-\bm{Y}_{ik}|^{p-1}\sign(\bm{X}_{ik}-\bm{Y}_{ik})\bm{G}_{ik} + \delta \\
    &=s(f_p, f_c) -\frac{1}{NK}\sum_{i=1}^N\sum_{k=1}^K p|\bm{X}_{ik}-\bm{Y}_{ik}|^{p-1}\sign(\bm{X}_{ik}-\bm{Y}_{ik})\nabla_{\bm{\theta}_p}\bm{X}_{ik}(\bm{\theta}_c-\bm{\theta}_p) + \delta \\
    &=s(f_p, f_c) -\frac{p}{NK}\nabla_{\bm{\theta}_p}\big[\sum_{i=1}^N\sum_{k=1}^K \sg[|\bm{X}_{ik}-\bm{Y}_{ik}|^{p-1}\sign(\bm{X}_{ik}-\bm{Y}_{ik})]\bm{X}_{ik}\big](\bm{\theta}_c-\bm{\theta}_p) + \delta \\
    &=s(f_p, f_c) -\frac{p}{NK}\nabla_{\bm{\theta}_p}\langle\sg[|\bm{X}-\bm{Y}|^{p-1}\sign(\bm{X}-\bm{Y})],\bm{X}\rangle(\bm{\theta}_c-\bm{\theta}_p) + \delta
\end{align}
\subsection{Linear Approximation of $log-sum-exp$ Function}
\begin{align}
    s(\Bar{f}_p, f_c)&=-\frac{1}{NKt}\sum_{i=1}^N\log\sum_{k=1}^K e^{t|\bm{\Bar{X}}_{ik}-\bm{Y}_{ik}|}\\
    &=-\frac{1}{NKt}\sum_{i=1}^N\log\sum_{k=1}^K e^{t|\bm{X}_{ik}+\bm{G}_{ik}-\bm{Y}_{ik}|} \\
    &\approx-\frac{1}{NKt}\sum_{i=1}^N\log\sum_{k=1}^K e^{t|\bm{X}_{ik}-\bm{Y}_{ik}|}-\frac{1}{NK}\sum_{i=1}^N\sum_{k=1}^K\frac{\bm{G}_{ik}e^{t|\bm{X}_{ik}-\bm{Y}_{ik}|}\sign(\bm{X}_{ik}-\bm{Y}_{ik})}{\sum_{j=1}^K e^{t|\bm{X}_{ij}-\bm{Y}_{ij}|}} + \delta \\
    &=s(f_p, f_c) -\frac{1}{NK}\sum_{i=1}^N\sum_{k=1}^K\frac{\bm{G}_{ik}e^{t|\bm{X}_{ik}-\bm{Y}_{ik}|}\sign(\bm{X}_{ik}-\bm{Y}_{ik})}{\sum_{j=1}^K e^{t|\bm{X}_{ij}-\bm{Y}_{ij}|}} + \delta \\
    &=s(f_p, f_c) -\frac{1}{NK}\sum_{i=1}^N\sum_{k=1}^K[\frac{e^{t|\bm{X}_{ik}-\bm{Y}_{ik}|}\sign(\bm{X}_{ik}-\bm{Y}_{ik})}{\sum_{j=1}^K e^{t|\bm{X}_{ij}-\bm{Y}_{ij}|}}\nabla_{\bm{\theta}_p}\bm{X}_{ik}(\bm{\theta}_c-\bm{\theta}_p)] + \delta \\
    &=s(f_p, f_c) -\frac{1}{NK}\nabla_{\bm{\theta}_p}\langle\sg[\softmax(t|\bm{X}-\bm{Y}|)*\sign(\bm{X}-\bm{Y})],\bm{X}\rangle(\bm{\theta}_c-\bm{\theta}_p) + \delta
\end{align}
\subsection{Linear Approximation of Centered Kernel Alignment}
The calculation of the squared CKA with the linear kernel can be written as 
\begin{align}
    s(\Bar{f}_p, f_c) = \frac{\langle \bm{H}\Bar{\bm{X}}\Bar{\bm{X}}^T\bm{H},\bm{H}\bm{Y}\bm{Y}^T\bm{H}\rangle^2}{\langle \bm{H}\Bar{\bm{X}}\Bar{\bm{X}}^T\bm{H},\bm{H}\Bar{\bm{X}}\Bar{\bm{X}}^T\bm{H}\rangle\langle \bm{H}\bm{Y}\bm{Y}^T\bm{H},\bm{H}\bm{Y}\bm{Y}^T\bm{H}\rangle},
\end{align}
which can be evaluated step by step after substituting the definition of $\Bar{\bm{X}}$ into it,
\begin{align}
    \Bar{\bm{X}}\Bar{\bm{X}}^T &= \bm{X}\bm{X}^T +\bm{G}\bm{X}^T+\bm{X}\bm{G}^T+\delta,\\
    \bm{H}\Bar{\bm{X}}\Bar{\bm{X}}^T\bm{H} &= \bm{H}\bm{X}\bm{X}^T\bm{H} + \bm{H}(\bm{G}\bm{X}^T+\bm{X}\bm{G}^T)\bm{H} +\delta,\\
    \langle \bm{H}\bm{X}\bm{X}^T\bm{H},\bm{H}(\bm{G}\bm{X}^T+\bm{X}\bm{G}^T)\bm{H}\rangle &=2\sum_{i=1}^N [\nabla_{\bm{\theta}_p}f_p(\bm{x}^{(i)})(\bm{\theta}_c-\bm{\theta}_p)]^T(\bm{X}-\frac{1}{N}\bm{1}\bm{1}^T\bm{X})^T(\bm{H}_{i:}\bm{X}\bm{X}^T\bm{H}-\frac{1}{N}\bm{1}^T\bm{H}\bm{X}\bm{X}^T\bm{H})^T\\
    &=\nabla_{\bm{\theta}_p}\Big[2\sum_{i=1}^N\big[\sg[(\bm{H}_{i:}\bm{X}\bm{X}^T\bm{H}-\frac{1}{N}\bm{1}^T\bm{H}\bm{X}\bm{X}^T\bm{H})(\bm{X}-\frac{1}{N}\bm{1}\bm{1}^T\bm{X})] f_p(\bm{x}^{(i)}\big]\Big](\bm{\theta}_c-\bm{\theta}_p),\\
    \langle \bm{H}\bm{Y}\bm{Y}^T\bm{H},\bm{H}(\bm{G}\bm{X}^T+\bm{X}\bm{G}^T)\bm{H}\rangle &=2\sum_{i=1}^N [\nabla_{\bm{\theta}_p}f_p(\bm{x}^{(i)})(\bm{\theta}_c-\bm{\theta}_p)]^T(\bm{X}-\frac{1}{N}\bm{1}\bm{1}^T\bm{X})^T(\bm{H}_{i:}\bm{Y}\bm{Y}^T\bm{H}-\frac{1}{N}\bm{1}^T\bm{H}\bm{Y}\bm{Y}^T\bm{H})^T\\
    &=\nabla_{\bm{\theta}_p}\Big[2\sum_{i=1}^N\big[\sg[(\bm{H}_{i:}\bm{Y}\bm{Y}^T\bm{H}-\frac{1}{N}\bm{1}^T\bm{H}\bm{Y}\bm{Y}^T\bm{H})(\bm{X}-\frac{1}{N}\bm{1}\bm{1}^T\bm{X})] f_p(\bm{x}^{(i)}\big]\Big](\bm{\theta}_c-\bm{\theta}_p).
\end{align}
The similarity can be approximated by Taylor expansion as 
\begin{align}
        s(\Bar{f}_p, f_c) &= s(f_p, f_c)(1 + 2\frac{\langle \bm{H}\bm{Y}\bm{Y}^T\bm{H},\bm{H}(\bm{G}\bm{X}^T+\bm{X}\bm{G}^T)\bm{H}\rangle}{\langle\bm{H}\bm{X}\bm{X}^T\bm{H},\bm{H}\bm{Y}\bm{Y}^T\bm{H}\rangle}-2\frac{\langle \bm{H}\bm{X}\bm{X}^T\bm{H},\bm{H}(\bm{G}\bm{X}^T+\bm{X}\bm{G}^T)\bm{H}\rangle}{\langle\bm{H}\bm{X}\bm{X}^T\bm{H},\bm{H}\bm{X}\bm{X}^T\bm{H}\rangle})+\delta\\
        &= s(f_p, f_c)\big[1 + \nabla_{\bm{\theta}_p}[\sum_{i=1}^N\sg(\zeta_i)f_p(\bm{x}^{(i)})](\bm{\theta}_c-\bm{\theta}_p)\big]+\delta,
\end{align}
where 
\begin{align}
    \zeta_i &= 4\frac{(\bm{H}_{i:}\bm{Y}\bm{Y}^T\bm{H}-\frac{1}{N}\bm{1}^T\bm{H}\bm{Y}\bm{Y}^T\bm{H})(\bm{X}-\frac{1}{N}\bm{1}\bm{1}^T\bm{X})}{\langle\bm{H}\bm{X}\bm{X}^T\bm{H},\bm{H}\bm{Y}\bm{Y}^T\bm{H}\rangle}\nonumber\\
    &\quad-4\frac{(\bm{H}_{i:}\bm{X}\bm{X}^T\bm{H}-\frac{1}{N}\bm{1}^T\bm{H}\bm{X}\bm{X}^T\bm{H})(\bm{X}-\frac{1}{N}\bm{1}\bm{1}^T\bm{X})}{\langle\bm{H}\bm{X}\bm{X}^T\bm{H},\bm{H}\bm{X}\bm{X}^T\bm{H}\rangle}\\
    &= 4\frac{\bm{H}_{i:}\bm{Y}\bm{Y}^T\bm{H}\bm{X}}{\langle\bm{H}\bm{X}\bm{X}^T\bm{H},\bm{H}\bm{Y}\bm{Y}^T\bm{H}\rangle}
    -4\frac{\bm{H}_{i:}\bm{X}\bm{X}^T\bm{H}\bm{X}}{\langle\bm{H}\bm{X}\bm{X}^T\bm{H},\bm{H}\bm{X}\bm{X}^T\bm{H}\rangle}
    .
\end{align}
\subsection{Linear Approximation of Distance Correlation}
The calculation of Distance Correlation can be written as
\begin{align}
    s(\Bar{f}_p, f_c) &= \frac{\langle \bm{H}\bm{D}_{\bar{\bm{X}}}\bm{H},\bm{H}\bm{D}_{\bm{Y}}\bm{H}\rangle^2}{\langle \bm{H}\bm{D}_{\bar{\bm{X}}}\bm{H},\bm{H}\bm{D}_{\bar{\bm{X}}}\bm{H}\rangle\langle \bm{H}\bm{D}_{\bm{Y}}\bm{H},\bm{H}\bm{D}_{\bm{Y}}\bm{H}\rangle},
\end{align}
where the distance matrices $\bm{D}_{\bar{\bm{X}}}$ and $\bm{D}_{\bm{Y}}$ are defined as following:
\begin{align}
    (\bm{D}_{\bm{X}})_{ij} &= ||\bm{X}_{i:}-\bm{X}_{j:}||_2,\\
    (\bm{D}_{\bm{Y}})_{ij} &= ||\bm{Y}_{i:}-\bm{Y}_{j:}||_2,\\
    \bm{D}_{ij:} &= \frac{\bm{X}_{i:}-\bm{X}_{j:}}{||\bm{X}_{i:}-\bm{X}_{j:}||_2},\\
    (\bm{D}_{\bar{\bm{X}}})_{ij} &= ||\bar{\bm{X}}_{i:}-\bar{\bm{X}}_{j:}||_2 = (\bm{D}_{\bm{X}})_{ij} + \bm{D}_{ij:}(\bm{G}_{i:}-\bm{G}_{j:})
\end{align}
Note that the $\bm{D}\in\mathbb{R}^{N\times N\times K}$ defined above is a third-order tensor, whose indexing is similar to the previously used one for matrix. By substituting the definitions above into the Distance Correlation, its approximation can be obtained:
\begin{align}
    s(\Bar{f}_p, f_c) &= s(f_p, f_c)\Big[1 + \frac{4}{\langle \bm{H}\bm{D}_{\bm{X}}\bm{H},\bm{H}\bm{D}_{\bm{Y}}\bm{H}\rangle} \sum_{i=1}^N\Big[\sum_{j=1}^N\big[[(\bm{H}\bm{D}_{\bm{Y}}\bm{H})_{ij}-\frac{1}{N}\sum_{q=1}^N (\bm{H}\bm{D}_{\bm{Y}}\bm{H})_{jq}](\bm{D}_{ij:}-\frac{1}{N}\sum_{q=1}\bm{D}_{iq:})\big]\bm{G}_{i:}^T\Big]
    \nonumber\\ 
    &\quad- \frac{4}{\langle \bm{H}\bm{D}_{\bm{X}}\bm{H},\bm{H}\bm{D}_{\bm{X}}\bm{H}\rangle}\sum_{i=1}^N\Big[\sum_{j=1}^N\big[[(\bm{H}\bm{D}_{\bm{X}}\bm{H})_{ij}-\frac{1}{N}\sum_{q=1}^N (\bm{H}\bm{D}_{\bm{X}}\bm{H})_{jq}](\bm{D}_{ij:}-\frac{1}{N}\sum_{q=1}\bm{D}_{iq:})\big]\bm{G}_{i:}^T\Big]\Big]+\delta\\
    &= s(f_p, f_c)\big[1 + \nabla_{\bm{\theta}_p}[\sum_{i=1}^N\sg(\xi_i)f_p(\bm{x}^{(i)})](\bm{\theta}_c-\bm{\theta}_p)\big]+\delta,
\end{align}
where
\begin{align}
    \xi_i&= \frac{4\sum_{j=1}^N\big[[(\bm{H}\bm{D}_{\bm{Y}}\bm{H})_{ij}-\frac{1}{N}\sum_{q=1}^N (\bm{H}\bm{D}_{\bm{Y}}\bm{H})_{jq}](\bm{D}_{ij:}-\frac{1}{N}\sum_{q=1}\bm{D}_{iq:})\big]}{\langle \bm{H}\bm{D}_{\bm{X}}\bm{H},\bm{H}\bm{D}_{\bm{Y}}\bm{H}\rangle} \nonumber \\
    &\quad - \frac{4\sum_{j=1}^N\big[[(\bm{H}\bm{D}_{\bm{X}}\bm{H})_{ij}-\frac{1}{N}\sum_{q=1}^N (\bm{H}\bm{D}_{\bm{X}}\bm{H})_{jq}](\bm{D}_{ij:}-\frac{1}{N}\sum_{q=1}\bm{D}_{iq:})\big]}{\langle \bm{H}\bm{D}_{\bm{X}}\bm{H},\bm{H}\bm{D}_{\bm{X}}\bm{H}\rangle}\\
    &= \frac{4\sum_{j=1}^N[(\bm{H}\bm{D}_{\bm{Y}}\bm{H})_{ij}\bm{D}_{ij:}]}{\langle \bm{H}\bm{D}_{\bm{X}}\bm{H},\bm{H}\bm{D}_{\bm{Y}}\bm{H}\rangle} - \frac{4\sum_{j=1}^N[(\bm{H}\bm{D}_{\bm{X}}\bm{H})_{ij}\bm{D}_{ij:}]}{\langle \bm{H}\bm{D}_{\bm{X}}\bm{H},\bm{H}\bm{D}_{\bm{X}}\bm{H}\rangle}\\
    &= \frac{4\bm{H}_{i:}\bm{D}_{\bm{Y}}\bm{H}\bm{D}_{i::}}{\langle \bm{H}\bm{D}_{\bm{X}}\bm{H},\bm{H}\bm{D}_{\bm{Y}}\bm{H}\rangle} - \frac{4\bm{H}_{i:}\bm{D}_{\bm{X}}\bm{H}\bm{D}_{i::}}{\langle \bm{H}\bm{D}_{\bm{X}}\bm{H},\bm{H}\bm{D}_{\bm{X}}\bm{H}\rangle}
\end{align}
\subsection{Connection Between Linearization and Minimal Distance}
In this subsection, we solve the two optimizations in Proposition 2 in the main text. For the first maximization problem, $\bm{W}(\bm{x})$ can be solved for each $\bm{W}$. 
\begin{align}
    \max_{\bm{W}(\bm{x})} s(f_p'(\bm{x}),f_c(\bm{x})) = \max_{\bm{W}(\bm{x})} -\frac{1}{K}||f'_p(\bm{x}) - f_c(\bm{x})||_2^2\max_{\bm{W}(\bm{x})} -\frac{1}{K}||f_p(\bm{x}) + \bm{W}(\bm{x})(\bm{\theta}_c - \bm{\theta}_p)- f_c(\bm{x})||_2^2.
\end{align}
Taking the derivative, the optimal condition is 
\begin{align}
    \bm{0} = (f_p(\bm{x}) + \bm{W}(\bm{x})(\bm{\theta}_c - \bm{\theta}_p)- f_c(\bm{x}))(\bm{\theta}_c - \bm{\theta}_p)^T.
\end{align}
Thus, 
\begin{align}
    \bm{W}(\bm{x}) = (f_c(\bm{x}) -f_p(\bm{x}))(\bm{\theta}_c - \bm{\theta}_p)^T[(\bm{\theta}_c - \bm{\theta}_p)(\bm{\theta}_c - \bm{\theta}_p)^T]^{-1}.
\end{align}
Because $f_c$ and $f_p$ are real parent-child pair, $f_c$ can be approximated by the Linearized model, in other words, $f_c$ can be estimated by the Taylor expansion at $f_p$. Thus, 
\begin{align}
    \bm{W}(\bm{x}) &= \nabla_{\bm{\theta}_p}f_c(\bm{x})(\bm{\theta}_c - \bm{\theta}_p)(\bm{\theta}_c - \bm{\theta}_p)^T[(\bm{\theta}_c - \bm{\theta}_p)(\bm{\theta}_c - \bm{\theta}_p)^T]^{-1} + \delta \approx \nabla_{\bm{\theta}_p}f_c(\bm{x}).
\end{align}
Similarly, the optimal condition for the second optimization is 
\begin{align}
    \sum_{i=1}^N[\nabla_{\bm{\theta}_p}f_p(\bm{x}^{(i)})]^T\nabla_{\bm{\theta}_p}f_p(\bm{x}^{(i)})\bm{Z} = \sum_{i=1}^N [\nabla_{\bm{\theta}_p}f_c(\bm{x}^{(i)})]^T[f_c(\bm{x}^{(i)})-f_p(\bm{x}^{(i)})]
\end{align}
By Taylor expansion, the right-hand side can be rewritten as  $\sum_{i=1}^N[\nabla_{\bm{\theta}_p}f_p(\bm{x}^{(i)})]^T\nabla_{\bm{\theta}_p}f_p(\bm{x}^{(i)})(\bm{\theta}_c - \bm{\theta}_p)$. Thus, $\bm{Z} =(\bm{\theta}_c - \bm{\theta}_p)$.

\section{Further Discussions}
\subsection{The Applicability of Linearized Network}
Our approach does not align perfectly with the NTK setting, as the NTK method employs Gaussian initialization for the neural network parameters. In contrast, our model's initialization stems from a pre-training phase. Given that \cite{ntk} demonstrated that the influence of gradient descent on individual parameter values is minimal, and it primarily affects the collective behavior of parameters, i.e., the features and outputs of the neural network. Consequently, we can continue to leverage the relevant results from the NTK framework.
\subsection{Potential Application}
Lineage detection identifies connections between child models and ancestors, enabling the detection/classification of inherited failure modes. For example, given some parent models and a child model, suppose each parent has a certain failure mode, lineage detection identifies the child's most probable parent or determines the probability of belonging to each parent, thereby assessing which failure mode the child is likely to inherit or the probability of inheriting various failure modes. The child model can then be cleansed using unlearning or debiasing methods. In fact, this process can also be reversed: upon observing certain failure modes in a child model, lineage detection probabilities can be used to evaluate the likelihood of each parent model having these failure modes in a Bayesian way, leading to targeted cleansing of the parent models.
\subsection{Limitations and Future Explorations}
As the very first exploration in the thread of model lineage detection, we have endeavored to encompass a diverse range of computer vision tasks, including classification, segmentation, and detection. We have delved into various learning paradigms, from direct fine-tuning and multi-stage fine-tuning to continual learning and knowledge distillation. However, numerous settings and deep learning tasks remain outside the scope of this paper. For instance, when dealing with dynamic neural networks, the challenge arises of concurrently addressing lineage prediction and the variability in model parameter space dimensions. This limitation primarily arises from our method's requirement that the parent model and child models be naturally aligned. We leave lineage detection with architecture modifications as future research work. Furthermore, although our lineage detector can also be applied in settings that only utilize neural network outputs, the task we addressed in this work is more akin to a white-box setup, if applying the taxonomy in the field of adversarial learning, requiring knowledge of both the network's parameters and its feature outputs. This setup is justified, as the finetuning process inherently involves complex interactions between parameters and output features.
Indeed, the prevalence of models developed based on open-source projects also presents a viable application scenario for our white-box approach to lineage detection. 
However, compared to a black-box setup, our task can be relatively easier. As the very first work for lineage detection, investigating this ideal white-box scenario is indispensable. We hope our work will contribute to a deeper understanding of lineage detection tasks and serve as a foundation for future research. We leave the exploration of black-box lineage detection and other more challenging lineage detection tasks for future works.

\end{document}